\pdfoutput=1 
\documentclass{article}
\usepackage{iclr2027_conference,times}
\usepackage{hyperref}
\usepackage{url}
\usepackage{graphicx} 
\usepackage{tikz} 
\usetikzlibrary{patterns}
\usepackage{pgfplots}
\pgfplotsset{compat=1.18}
\usepackage{algorithm}   
\usepackage{algorithmic} 
\usepackage{listings}
\usepackage[skins,breakable]{tcolorbox}

\definecolor{cNoTh}{HTML}{C9CDD3}
\definecolor{cPlanAct}{HTML}{AAB1BA}
\definecolor{cReActC}{HTML}{8B939D}
\definecolor{cReAct}{HTML}{C9CDD3}
\definecolor{cReflAct}{HTML}{6B7280}
\definecolor{cTeal}{HTML}{0F9B8E}

\newcommand{\method}{SGG-ReflAct}
\newcommand{\beammethod}{BeamSGG-ReflAct}
\newcommand{\reflact}{\textsc{ReflAct}}
\newcommand{\react}{\textsc{ReAct}}

\newcommand{\text}[1]{\mbox{#1}}
\newcommand{\boldsymbol}[1]{\mathbf{#1}}
\newcommand{\mathds}[1]{\mathrm{#1}}

\newcommand{\toprule}{\hline}
\newcommand{\midrule}{\hline}
\newcommand{\bottomrule}{\hline}
\newcommand{\ourrow}{}
\newcommand{\multirow}[3]{#3}

\newcommand{\checkmark}{\ensuremath{\surd}}

\newtcolorbox{promptbox}[1]{
  enhanced, breakable,
  colback=white, colframe=black!60, boxrule=0.5pt, arc=1pt,
  colbacktitle=black!8, coltitle=black, fonttitle=\bfseries\small,
  title={#1},
  left=6pt, right=6pt, top=3pt, bottom=4pt,
}

\title{SGG-ReflAct: Sub-Goal Guided ReflAct\\
with Structured Planning\\
for Reliable Long-Horizon Reasoning}

\author{Jaeho Jung \\
Independent Researcher \\
Seoul, Republic of Korea \\
\texttt{gkgkgkwogh@gmail.com} \\
\And
Sung Hoon Jung \thanks{Corresponding author.} \\
Department of Applied Artificial Intelligence \\
Hansung University \\
Seoul, Republic of Korea \\
\texttt{shjung@hansung.ac.kr}
}

\iclrfinalcopy 

\begin{document}

\maketitle

\begin{abstract}
Recent advances in reasoning backbones have empowered large language model (LLM)
agents to tackle complex, multi-step tasks. However, as reasoning horizons grow,
inconsistent internal beliefs induce intermediate errors that cause agents to
drift from their goals. This limitation also persists in \reflact{}, which
reflects only on the end-goal at each step without explicitly considering
intermediate sub-goals. To address this problem, we propose \method{}
(Sub-Goal Guided ReflAct), a reasoning backbone that integrates sub-goals
generated through a single-path LLM planner into the reflection process. We
further extend this framework to \beammethod{}, which replaces the single-path
planner with a beam search-based LLM planner for structured plan exploration.
We run experiments on ALFWorld, ScienceWorld, and Jericho with multiple LLM
models. \method{} outperforms \reflact{} in nearly all settings, achieving best success rate gains of 14.9 percentage points on
ALFWorld and 8.0 percentage points on ScienceWorld with
Llama-3.1-8B-Instruct. Our experimental analysis shows that \method{} reduces hallucinated actions and achieves its
largest gains on procedurally ordered tasks. Furthermore, experimental results
with \beammethod{} show that the
backbone's effectiveness depends on plan quality:
explicitly specifying the required operations recovers gains that plan searching
alone cannot achieve. These results demonstrate that \method{} offers a
practical and highly effective reasoning backbone, enabling LLM agents to
achieve reliable performance in complex, long-horizon tasks through easy
integration.
\end{abstract}

\begin{figure}[t]
\centering
\includegraphics[width=0.63\linewidth]{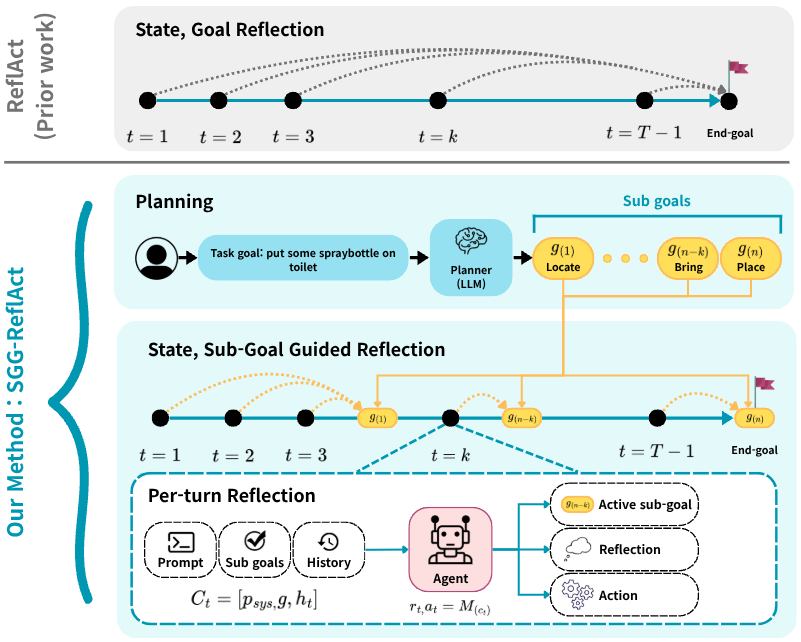}
\caption{
  Overview of \method{} on an ALFWorld task. (Top) \reflact{} anchors per-turn
  reflections to a fixed end-goal. (Bottom) \method{} generates a sub-goal
  plan $\mathcal{G} = (g_1, \ldots, g_N)$ via a single planning call,
  anchoring reflections to the active sub-goal to allow the target to
  move with the agent. The inset illustrates the inference at turn $t = k$, where
  context $c_t$ (system prompt, sub-goal plan, and history) is used to predict the
  active
  sub-goal, reflection $r_t$, and next action $a_t$ in a single step. \method{}
  incurs minimal overhead, adding only one initial planning call.
}
\label{fig:overview}
\end{figure}

\section{Introduction}
\label{sec:introduction}

Recent advances in Large Language Models (LLMs) have expanded their capabilities, enabling LLM-based agents to tackle complex tasks that require more than a single response \citep{brown2020gpt3, wang2024survey}.
Such tasks demand multi-step decision making under partial observability and
sparse rewards, so the agent has to reason, act, and observe over a long horizon
\citep{yao2023react, shridhar2020alfworld, wang2022scienceworld}.
A \emph{reasoning backbone} governs that loop by fixing how the agent thinks
before it acts, and its design has therefore become a central focus of LLM agent
research.

Existing research follows three primary directions. The first, such as \react{}, interleaves reasoning with action, anchoring each thought to the most recent observation \citep{wei2022cot, yao2023react}. The second, such as Plan-and-Act, provides explicit structure through planning and searching over future trajectories \citep{jiang2023plan, erdogan2025planact, mpo2025, zhou2024lats}. The third employs reflection on top of a reasoning backbone, revisiting failed episodes to inform subsequent attempts \citep{shinn2023reflexion}.

Despite these diverse paradigms, all three share a fundamental vulnerability in long-horizon interactions: ungrounded steps enter unchecked and compound \citep{reflact2025, xie2023selfevalguided}, one misaligned thought triggering a hallucinated action that yields a misleading observation and corrupts the next thought, so the agent's belief drifts from the environment until reasoning collapses.
\reflact{} \citep{reflact2025} attacks this at its source, redesigning the
reasoning--action cycle itself: at each timestep the agent weighs its current
state against the task goal before action selection, which grounds reasoning in
the environment state. This works well, but weakens in long-horizon tasks, when
early in an episode the state is still far from the end-goal and reflecting on
it offers little concrete guidance toward the next step. We therefore conjecture
that anchoring each reflection to a nearer, intermediate target rather than the
distant end-goal gives more actionable guidance while still far from
completion. Our proposed anchoring directly addresses this issue, preventing
early misjudgments from compounding into the ungrounded actions examined in
Section~\ref{sec:task_analysis}.

Motivated by this, we propose \textbf{\method{}} (\textbf{S}ub-\textbf{G}oal
\textbf{G}uided \textbf{ReflAct}), a reasoning backbone that bridges structured
planning and reflective acting (Figure~\ref{fig:overview}).
The mechanism is twofold.
First, a \textit{Planner} module runs once at episode start and decomposes the
task goal into a sub-goal plan, each sub-goal serving as an explicit milestone.
Second, at each timestep the agent weighs its present state against \emph{the active sub-goal}, rather than the
distant end-goal, before action selection; the reflection target thus advances with
the agent and stays proximate as the horizon grows.
Due to the modular nature of the planner, we can isolate the impact of search
effort by keeping the reflection-and-acting phase constant. This allows us to
evaluate whether increased computation on plan search yields better
performance than direct planning. Specifically, we introduce
\textbf{\beammethod{}}, which replaces the single-path planner with a
self-evaluation guided beam search over candidate sub-goal sequences
\citep{xie2023selfevalguided}.

We evaluate both methods on three models (GPT-4o, GPT-4o-mini, and
Llama-3.1-8B-Instruct) and three interactive text benchmarks: ALFWorld
\citep{shridhar2020alfworld}, ScienceWorld \citep{wang2022scienceworld}, and
Jericho \citep{hausknecht2020jericho}.
\method{} outperforms the \react{}, Plan-and-Act, and \reflact{} baselines in
nearly all settings.
The gains are largest on the resource-constrained Llama-3.1-8B-Instruct, where
\method{} reaches 50.7\% on ALFWorld and 33.6\% on ScienceWorld.
These results surpass \reflact{} by +14.9\,pp and +8.0\,pp.
Our work makes the following contributions:
\begin{enumerate}
    \item We propose \method{}, a reasoning backbone that bridges sub-goal
        planning and per-turn reflection. Each reflection is anchored to the
        active sub-goal rather than the distant end-goal, keeping the reflection
        target proximate as the horizon grows; its planner is a \emph{modular}
        component, so a different planner can be swapped in without altering the
        reflection-and-acting backbone, and we instantiate this with
        \beammethod{}.

    \item Through a \emph{grounding} analysis we show that the gains stem not
        from merely adding plans to the context---as previously noted by
        \citet{reflact2025}---but from anchoring each reflection to the active
        sub-goal. This mitigates the accumulation of hallucinated actions during
        long-horizon episodes, reducing the invalid-action rate from
        43.3\% to 29.9\% on Llama-3.1-8B-Instruct.

    \item Through a \emph{plan-content} analysis we show that plan quality
        determines the backbone's effectiveness; inadequate plans can turn gains
        into degradation, and successful plans must specify the operations
        required to achieve the goal rather than merely restating it.
        Explicitly prompting the planner to include these operations recovers
        gains that plan search alone cannot achieve.
\end{enumerate}

\section{Related Work}
\label{sec:related}

\paragraph{LLM agent reasoning backbones.}

Structured reasoning has been at the center of progress in LLM agents:
Chain-of-Thought (CoT) prompting \citep{wei2022cot} first showed that reasoning
step by step improves answers on hard tasks, and set the stage for agent
frameworks. \react{} \citep{yao2023react} took the next step, interleaving
reasoning and action so each thought is tied to the latest observation.
\react{}, however, forms each thought from the current observation and the past
history alone, without tying it to the end-goal or to how much of the task is
done; this gap becomes costly as the chain grows, because early mistakes carry
forward and build into a full breakdown \citep{xie2023selfevalguided, reflact2025}.

Several methods add modules on top of \react{}: WKM \citep{qiao2024wkm} and
KnowAgent \citep{zhu2025knowagent} keep a world model and an action knowledge
base to rule out infeasible moves, while MPO \citep{mpo2025} and Dynamic
Cheatsheet \citep{suzgun2025dynamic} hand the agent an optimized meta-plan or an
adaptive memory, and Trial and Error \citep{song2024trialerror} searches the
action space directly. All add knowledge, memory, or a plan, but leave the core
thought-action loop unchanged, so they help little when the backbone itself
produces ungrounded thoughts.

\paragraph{Agent planning and search.}

A second line builds a plan before or during execution. Plan-and-Solve
\citep{jiang2023plan} and Plan-and-Act \citep{erdogan2025planact} turn the task
into a high-level plan at the start and run it step by step, giving a structure
that plain \react{} lacks; Pre-Act \citep{rawat2025preact} instead keeps updating
the plan as each step returns new output, and SwiftSage \citep{swiftsage2023}
uses a hybrid approach. Like Plan-and-Solve and Plan-and-Act, we fix the plan for the whole episode;
what differs is that we use it as a target for reflection rather than a script to
execute, so the extra cost stays at one planning call. Others organize the plan
itself---ReAcTree \citep{choi2026reactree} as a control-flow tree, PlanU
\citep{planu2025} under uncertainty over sub-goal orderings, and plans refined
from retrieved experience and graph-based checks
\citep{kagaya2024rap, jeong2026tape}---while RAP \citep{hao2023rap} plans with a
world model and RAFA \citep{rafa2024} and LATS \citep{zhou2024lats} reason over
trajectories or rollouts. Most of these plan-centric methods stay with their
plan and act without per-turn state--goal reflection, so they turn fragile when
an observation does not fit the plan.

Search-based methods explore several reasoning paths at inference time:
Tree-of-Thoughts (ToT) \citep{yao2023tot} and Graph-of-Thoughts (GoT)
\citep{besta2024got} grow CoT reasoning into a branching search, and
Self-Evaluation Guided Beam Search \citep{xie2023selfevalguided} runs beam search
over candidates scored by the LLM itself, and rStar \citep{qi2024rstar} lets two
LLMs reason against each other to steer the search. These work well for static or
single-step reasoning, but do not carry over directly to sequential, partially
observable environments. \beammethod{} adapts self-evaluation guided beam search
to plan building instead, scoring sub-goal sequences by feasibility and order
rather than answer correctness, so the planner stays a drop-in part of
\method{}.

\paragraph{Reflection mechanisms.}

A growing line of work has agents reflect on and fix their own behavior.
Reflexion \citep{shinn2023reflexion} writes a verbal self-reflection after a
failed episode, Self-Refine \citep{madaan2023selfrefine} and CRITIC
\citep{gou2024critic} revise an output using self- or tool-generated feedback,
and ExpeL \citep{zhao2024expel} distills reusable lessons from past runs; all
reflect on \emph{outputs} or \emph{past episodes}, outside the per-turn loop where
the errors first appear. Closer to that loop, ReBACT \citep{zeng2025rebact} fixes
errors before each action, and InferAct \citep{fang2025inferact} checks a
proposed action before it runs; both act as a \emph{verifier} placed after the
action. \reflact{} \citep{reflact2025} instead redesigns the backbone itself, so that at
each step the agent weighs its current state against the task goal rather than
just predicting the next action. Grounding each turn cuts belief mismatch sharply,
yet \reflact{}'s reflection still gives little help early in a long task, when
the current state is far from the end-goal. \method{} builds on this line,
keeping per-turn reflection but anchoring it to the active sub-goal instead of the
distant end-goal, so it differs from plan-injection methods such as MPO
\citep{mpo2025}, which add a plan to the context and leave the reflection
backbone unchanged (Table~\ref{tab:comparison}, Appendix~\ref{app:comparison_sec}).
This is consistent with \citet{reflact2025}, who found that adding MPO
meta-plans to a reflection backbone yields negligible performance gains, as a
plan is effective only when it redirects the per-turn reflection rather than
merely filling the context.

\section{Method}
\label{sec:method}

We formalize the task as a POMDP \citep{kaelbling1998planning}, following the
notation of \citet{reflact2025}. Let $u$ denote the task instruction, including
the end-goal. At each step $t$, the agent receives an observation $o_t$,
maintains a history $h_t = \{(o_1, a_1), \ldots, (o_{t-1}, a_{t-1}), o_t\}$, and
selects an action $a_t$ via a language model $\mathcal{M}$. Given the partial
observability of the environment, the agent must act on a belief formed from
$h_t$; errors in that belief lead to hallucinated actions. A reasoning backbone,
following \citet{yao2023react}, adds a natural-language thought $\tau_t$ before
$a_t$ to reduce them.

All three backbones receive the task instruction $u$, but differ in what the
per-turn instruction directs $\tau_t$ at. \react{} attends to the history alone,
$\tau_t,\, a_t = \mathcal{M}(h_t)$, with no link to the goal or to progress so far.
\reflact{} \citep{reflact2025} attends to the state relative to the goal,
producing a reflection $r_t^g$ toward the end-goal,
$\tau_t,\, r_t^g,\, a_t = \mathcal{M}(h_t)$, but in long-horizon tasks the
distance between the current state and end-goal in $u$ leaves the reflection loosely tied to
the immediate sub-task. To address this, \method{} attends to the active sub-goal
of a plan $\mathcal{G} = (g_1, \ldots, g_N)$, producing a reflection $r_t^{sg}$
toward that sub-goal,
$g^{*}_t,\, \tau_t,\, r_t^{sg},\, a_t = \mathcal{M}(h_t,\, \mathcal{G})$: at each
turn $t$ the agent identifies the \emph{active sub-goal} $g^{*}_t \in \mathcal{G}$
and reflects on it rather than the end-goal in $u$. This framework runs in two phases: a \emph{planning phase}
that decomposes the end-goal in $u$ into $\mathcal{G}$ once per episode, and a
\emph{reflection-and-acting phase} that runs every turn, doing sub-goal guided
reflection before acting. Figure~\ref{fig:overview} illustrates both phases and
contrasts our reflection target with that of \reflact{}.

\subsection{SGG-ReflAct Framework}
\label{sec:method:framework}

\paragraph{Phase 1: Goal Decomposition.}
The first phase is a dedicated \textit{Planner} module. It runs once at $t = 0$,
before the agent touches the environment, and maps an initial context $c_0$ and a
single in-context planning example $\mathcal{E}$ to an ordered sequence of $N$
high-level sub-goals:
\begin{equation}
  \mathcal{G} = \textsc{Planner}\!\left(c_0,\;\mathcal{E}\right)
              = \left(g_1,\, g_2,\, \ldots,\, g_N\right).
  \label{eq:planning}
\end{equation}
We refer to $\mathcal{G}$ simply as \emph{the plan}. The initial condition $c_0$
varies by environment: $c_0 = u$ for ScienceWorld,
and $c_0 = \{u, o_0\}$ for ALFWorld and Jericho to provide spatial context.
Each sub-goal $g_i$ defines a high-level objective (\emph{what to do}) rather than a
specific action (\emph{how to do}), using generalized phrasing to preserve plan
generality. These sub-goals are ordered by environmental prerequisites. We set
$N=4$ based on ScienceWorld's median sub-goal sequence length; to maintain one
unified architecture rather than tune $N$ per environment, we apply the same
value to ALFWorld and Jericho, where tasks also exhibit a comparable
four-stage structure (Appendices~\ref{app:nchoice}, \ref{app:planner_prompts},
\ref{app:sci_planner}, \ref{app:jericho_planner}). Producing this fixed
sequence $\mathcal{G}$ costs one additional LLM call per episode; because the
planner commits to a single candidate without search or revision, we call it
a \emph{single-path} planner. These sub-goals remain abstract and
\emph{phase-level} across all benchmarks to emphasize agent interaction.

\paragraph{Phase 2: Sub-Goal Guided Reflection and Acting.}
At every turn $t \geq 1$, the agent sees the plan $\mathcal{G}$ and the
history $h_t = \{(o_1, a_1), \ldots, (o_{t-1}, a_{t-1}), o_t\}$, with the plan
fixed at the front of the context:
\begin{equation}
  c_t = \Bigl[\,
    \underbrace{p_{\mathrm{sys}}}_{\text{system}},\;
    \underbrace{\mathcal{G}}_{\text{sub-goals}},\;
    \underbrace{h_t}_{\text{history}}
  \,\Bigr].
  \label{eq:context}
\end{equation}
The system prompt $p_{\mathrm{sys}}$ carries the task instruction $u$, so the
end-goal stays in the context throughout, and it asks the agent to reply in three
parts: the active sub-goal $g^{*}_t \in \mathcal{G}$ it is working on, a
one-sentence reflection on where it stands on $g^{*}_t$, and the next action
$a_t$.
Appendix~\ref{app:prompts} shows the full instruction and format.

\reflact{} tells the agent to reflect on its state against the end-goal; ours
instead has the agent first name the active sub-goal, then reflect on its state
against it. Since the agent can reach $g^{*}_t$ from its current state, while $u$
may remain many actions away, the reflection stays grounded in what the agent
can act on next, even early in a long episode. The agent then produces:
\begin{equation}
  g^{*}_t,\; \tau_t,\; r_t^{sg},\; a_t = \mathcal{M}\!\left(c_t\right),
  \label{eq:action}
\end{equation}
where $\mathcal{M}$ is the language model that runs the policy, $c_t$ is the
context from Equation~\ref{eq:context}, $g^{*}_t \in \mathcal{G}$ is the active
sub-goal the model selects for this turn, $\tau_t$ is its thought, $r_t^{sg}$ is
its reflection on $g^{*}_t$, and $a_t$ is the action. All four come from a
single call, so identifying the active sub-goal costs no extra LLM call.
Rather than four discrete fields, the model produces a single free-form response
where $g^{*}_t$ is described within the reflection; only $a_t$ is parsed via the
\texttt{Action:} tag.

\subsection{BeamSGG-ReflAct: Search-Enhanced Planning}
\label{sec:method:beam}

Since reflections are anchored to the active sub-goal from $\mathcal{G}$,
the agent's reasoning capability is inherently constrained by the plan's
coverage. Because the plan is static, any missing task requirements cannot be
addressed during the episode. To test whether searching the space of sub-goal
sequences improves plan quality, \beammethod{} replaces the single-path planner
with a \textit{Beam Search Planner}, which likewise runs once at $t=0$, before
the episode begins. This planner adapts self-evaluation guided beam search
\citep{xie2023selfevalguided} to construct $\mathcal{G}$ incrementally.
Unlike MCTS-based approaches such as LATS \citep{zhou2024lats}, which require
environment rollouts or world models, our planner relies exclusively on LLM
self-evaluation, requiring no additional environment interaction.

\paragraph{Beam Search Planner.}
The planner expands a beam of $K$ partial plans, adding one sub-goal at each
depth $d \in \{1, \dots, N\}$. At each step, it generates $S$ candidates per
beam, drawing them one at a time rather than all at once because decoding is
deterministic (Appendix~\ref{app:setup}): each new request is told which
candidates have already been produced at that depth and is asked for a
different one (see Appendix~\ref{app:beam}).
Writing $\boldsymbol{\rho} = (g_1, \ldots, g_d)$ for a candidate's partial
plan at depth $d$, candidates are scored by the product of their prefix
self-evaluation scores:
\begin{equation}
  \mathrm{score}(\boldsymbol{\rho}) = \prod_{i=1}^{|\boldsymbol{\rho}|}\,
    \textsc{SelfEval}\!\left(\boldsymbol{\rho}_{1:i} \;\middle|\; u\right),
  \label{eq:beam_score}
\end{equation}
then keeps the top $K$ for the next depth. Here $\textsc{SelfEval}(\cdot)$ is a
separate LLM call that rates a prefix on a $[0,1]$ scale against a fixed
rubric (Equation~\ref{eq:selfeval}). It judges the whole prefix rather than
the sub-goal just added, so an early sub-goal enters the product at every
later depth and carries more weight than a late one. For our experiments, we
set $N=4, K=2, S=3$. With two LLM calls per candidate (expansion and
evaluation), the search requires at most 42 calls per episode. Full details
are in Appendix~\ref{app:beam}.

\beammethod{} shares \method{}'s reflection-and-acting phase and differs only in
how $\mathcal{G}$ is built, so the gap between them measures the effect of the
sub-goal plan itself rather than making \beammethod{} a separate method; both
planner prompts impose the same constraints and differ only in the unit they
produce.

Setting $K=1$ and $S=6$ reduces the planner to a \emph{greedy} search that
retains one partial plan and samples six candidates per depth (up to 48 calls
per episode); Section~\ref{sec:plan_quality} evaluates both variants.

\section{Experiments}
\label{sec:experiments}

We ask whether sub-goal guided, per-turn reflection improves performance and
which of its two ingredients---the sub-goal plan and the per-turn
reflection---is responsible; the baselines therefore remove one at a time. The
benchmarks are long-horizon, where a reflection target far from the agent's
state has room to drift, and the models span three capacity levels, since
explicit sub-goals should matter most where the model has least to spare.

\subsection{Baselines}
Following \citet{reflact2025}, we compare against four frameworks chosen to span
the two ingredients of \method{}. \textbf{NoThinking} and \textbf{\react{}}
\citep{yao2023react} hold neither a sub-goal plan nor a state--goal reflection
and set the floor. \textbf{\reflact{}} \citep{reflact2025} supplies per-turn
reflection without sub-goals, so the reflection target alone separates it from
ours. \textbf{Plan-and-Act} \citep{erdogan2025planact}
supplies a first-timestep plan without per-turn reflection, its plan coming from
the agent's own first-turn thought rather than a dedicated planner, so what
separates it is what the agent does with the plan once the episode is under
way. A gain that
survives both single-ingredient baselines cannot be credited to either on its
own. Appendix~\ref{app:comparison_sec} places all of them on the same two axes.

\subsection{Benchmarks and Metrics}
We evaluate on three interactive text benchmarks that differ in domain and in
what they demand of a sub-goal plan.
\textbf{ALFWorld} \citep{shridhar2020alfworld} is a TextWorld
\citep{cote2019textworld} household environment whose standard 134-episode
split has six task types.
\textbf{ScienceWorld} \citep{wang2022scienceworld} covers procedural science
experiments in a 211-episode split of 24 task types (grouped into nine by the
split's identifiers); many enforce a strict prerequisite order, so a sub-goal
plan that names the required operation
pays off.
\textbf{Jericho} \citep{hausknecht2020jericho} is the hardest of the three: the
20-game AgentBoard \citep{ma2024agentboard} set pairs large action spaces with
sparse rewards, so a plausible-sounding sub-goal is often unreachable.
We report \textbf{Success Rate (SR)} across all benchmarks and \textbf{Average
Reward (AR)} for ScienceWorld and Jericho. Because the built-in success flags in
ScienceWorld and Jericho are unreliable, we follow \citet{reflact2025} and
derive success from progress rewards; Appendices~\ref{app:setup}
and~\ref{app:maxsteps} give the thresholds, the settings behind them, and the
per-task step budgets.

\subsection{Models and Implementation}
We evaluate on three models: \textbf{GPT-4o} and \textbf{GPT-4o-mini}
\citep{openai2024gpt4o} as proprietary models, and
\textbf{Llama-3.1-8B-Instruct} \citep{dubey2024llama} as a resource-constrained,
open-weight model. Llama-3.1-8B-Instruct is the most likely to benefit from
explicit sub-goals, so it is the primary subject of the analyses in
Sections~\ref{sec:task_analysis} and~\ref{sec:plan_quality}.
Appendix~\ref{app:setup} gives the decoding, prompting, and serving
settings, which are identical across agents.

\begin{table}[t!]
\centering
\caption{
  Performance comparison across benchmarks. Values are percentages. $\pm$
  denotes standard deviation across each benchmark's own task groupings:
  ALFWorld's six task types (Appendix~\ref{app:alfworld_cat}) and
  the ScienceWorld split's nine categories (Appendix~\ref{app:sciworld_cat}),
  whose
  tables report the same value in their \emph{SD} column. Jericho defines no
  such grouping, so none is reported for it. \textbf{Bold} indicates the best
  performance within each model group per column.
}
\label{tab:main}
\footnotesize
\setlength{\tabcolsep}{2.5pt}
\renewcommand{\arraystretch}{0.92}
\begin{tabular}{llccccc}
\toprule
\multirow{2}{*}{Model} &
\multirow{2}{*}{Agent} &
  \multicolumn{1}{c}{ALFWorld} &
  \multicolumn{2}{c}{ScienceWorld} &
  \multicolumn{2}{c}{Jericho} \\
\cline{3-3}\cline{4-5}\cline{6-7}
& & SR & SR & AR & SR & AR \\
\midrule
\multirow{6}{*}{GPT-4o}
& NoThinking                     & 78.4\,$\pm$\,16.8 & 53.6\,$\pm$\,29.4 & 66.6\,$\pm$\,21.5 & 20.0 & 45.8 \\
& \react{} \citep{yao2023react}  & 80.6\,$\pm$\,11.8 & 59.2\,$\pm$\,35.6 & 71.7\,$\pm$\,22.1 & 25.0 & 44.2 \\
& Plan-and-Act                   & 81.3\,$\pm$\,9.9 & 54.0\,$\pm$\,33.4 & 68.2\,$\pm$\,24.3 & 30.0 & 49.2 \\
& \reflact{} \citep{reflact2025} & 89.6\,$\pm$\,8.9 & 55.5\,$\pm$\,32.7 & 67.2\,$\pm$\,20.8 & 35.0 & 54.4 \\
\ourrow & \beammethod{} (ours)           & 84.3\,$\pm$\,14.7 & 63.5\,$\pm$\,31.5 & \textbf{74.9}\,$\pm$\,20.8 & 25.0 & 43.1 \\
\ourrow & \method{} (ours)               & \textbf{90.3}\,$\pm$\,8.9 & \textbf{64.0}\,$\pm$\,37.4 & 73.5\,$\pm$\,18.0 & \textbf{40.0} & \textbf{55.0} \\
\midrule
\multirow{6}{*}{GPT-4o-mini}
& NoThinking                     & 44.0\,$\pm$\,24.0 & 20.9\,$\pm$\,23.0 & 40.2\,$\pm$\,24.0 & 10.0 & 36.0 \\
& \react{} \citep{yao2023react}  & 45.5\,$\pm$\,17.4 & 40.8\,$\pm$\,33.0 & 53.2\,$\pm$\,20.7 & 5.0 & 26.1 \\
& Plan-and-Act                   & 43.3\,$\pm$\,9.8 & 36.0\,$\pm$\,33.4 & 51.0\,$\pm$\,28.0 & 5.0 & 21.1 \\
& \reflact{} \citep{reflact2025} & 62.7\,$\pm$\,21.3 & 40.3\,$\pm$\,34.5 & 55.6\,$\pm$\,23.9 & 15.0 & 34.3 \\
\ourrow & \beammethod{} (ours)           & 70.1\,$\pm$\,26.0 & 43.6\,$\pm$\,31.9 & 55.9\,$\pm$\,20.5 & 15.0 & 36.3 \\
\ourrow & \method{} (ours)               & \textbf{72.4}\,$\pm$\,21.3 & \textbf{45.0}\,$\pm$\,33.8 & \textbf{60.5}\,$\pm$\,22.8 & \textbf{20.0} & \textbf{38.6} \\
\midrule
\multirow{6}{*}{\shortstack[l]{Llama-3.1\\-8B-Instruct}}
& NoThinking                     & 28.4\,$\pm$\,25.2 & 13.3\,$\pm$\,16.7 & 34.6\,$\pm$\,19.5 & 5.0 & 26.3 \\
& \react{} \citep{yao2023react}  & 29.9\,$\pm$\,3.8 & 18.5\,$\pm$\,16.4 & 36.9\,$\pm$\,19.2 & 0.0 & 15.7 \\
& Plan-and-Act                   & 29.9\,$\pm$\,13.5 & 11.8\,$\pm$\,11.7 & 33.6\,$\pm$\,26.2 & 0.0 & 16.2 \\
& \reflact{} \citep{reflact2025} & 35.8\,$\pm$\,27.5 & 25.6\,$\pm$\,27.5 & 41.5\,$\pm$\,23.5 & 5.0 & \textbf{26.6} \\
\ourrow & \beammethod{} (ours)           & 46.3\,$\pm$\,22.5 & 26.5\,$\pm$\,24.0 & 41.9\,$\pm$\,23.3 & 5.0 & 17.8 \\
\ourrow & \method{} (ours)               & \textbf{50.7}\,$\pm$\,19.6 & \textbf{33.6}\,$\pm$\,26.2 & \textbf{48.7}\,$\pm$\,23.4 & \textbf{10.0} & 24.3 \\
\bottomrule
\end{tabular}
\renewcommand{\arraystretch}{1}
\end{table}

\subsection{Main Results}
\label{sec:results}

Table~\ref{tab:main} reports how \method{} and \beammethod{} compare with the
NoThinking, \react{}, Plan-and-Act, and \reflact{} baselines across the three
models and three benchmarks.


\paragraph{ALFWorld.}
\method{} leads in every model group, reaching 90.3\% with GPT-4o, 72.4\% with
GPT-4o-mini, and 50.7\% with Llama-3.1-8B-Instruct. On Llama this last number is
\textbf{+14.9\,pp} above \reflact{}. The single-path planner stays ahead of the
Beam Search Planner in every model group, by 2.3 to 6.0 points, and both are well
ahead of Plan-and-Act; planning and reflection thus help in different ways
(Section~\ref{sec:plan_quality}).

\paragraph{ScienceWorld.}
\method{} again leads every model group in SR, and in AR on
all but GPT-4o, where the Beam Search Planner edges ahead (74.9 against 73.5). On
Llama-3.1-8B-Instruct it improves over \reflact{} by \textbf{+8.0\,pp} SR and
\textbf{+7.2\,pp} AR.
A fixed plan without per-turn reflection can also hurt weaker models: Plan-and-Act drops
to 11.8\% on Llama, below even NoThinking (13.3\%), which agrees with
\citet{reflact2025}'s finding that smaller models struggle to follow a preset
plan.

\paragraph{Jericho.}
On the 20-game Jericho suite, \method{} has the best SR in every model group (for
example, 40.0\% against \reflact{}'s 35.0\% on GPT-4o) and the best AR on GPT-4o
and GPT-4o-mini. The Beam Search Planner, by contrast, never beats the single-path
planner on any model; on GPT-4o it falls to 25.0\%, behind even \reflact{} and
Plan-and-Act. We revisit this beam disadvantage across all three benchmarks in
Section~\ref{sec:plan_quality}.

\section{Discussion}
\label{sec:discussion}

The efficacy of \method{} is driven by two factors: \textbf{(C1)} the
necessity of reflection-target proximity for robust grounding
(Section~\ref{sec:task_analysis}) and \textbf{(C2)} the critical role of
plan content in determining backbone performance
(Section~\ref{sec:plan_quality}).

\subsection{ReflAct vs.\ SGG-ReflAct: Task-Level Success and Grounding (C1)}
\label{sec:task_analysis}

\paragraph{Impact of sub-goal plans vs.\ anchoring instructions.}
\method{} introduces two key modifications to \reflact{}: the inclusion of a
sub-goal plan in the context and the instruction to anchor each reflection to
the active sub-goal. To disentangle these effects, we conduct a controlled
ablation on ScienceWorld with Llama-3.1-8B-Instruct. \reflact{}+Plan uses the
same single-path planner as \method{} to generate its sub-goal plan, but
retains \reflact{}'s original instruction to reflect against the end-goal, so
the only deliberate difference between the two agents is what the per-turn
reflection targets (see
Appendix~\ref{app:ablation_example} for verbatim instructions). Merely
appending the plan to the context is not enough: without an instruction to
identify the currently active sub-goal, the agent has no mechanism for using it.

As shown in Table~\ref{tab:ablation} (Appendix~\ref{app:moved_tables}), of the +8.0\,pp improvement of \method{}
over \reflact{}, the mere presence of the plan in the context accounts for only
+1.9\,pp; the remaining +6.1\,pp (to 33.6\%) is attributed to anchoring the
reflection to the active sub-goal. Average reward follows a comparable trend:
the plan alone raises it by +3.3\,pp, and anchoring adds a further +3.9\,pp to
reach 48.7\%, aligning with \citet{reflact2025}, who found that adding a
meta-plan to a reflection backbone yields negligible improvements. Appendix~\ref{app:ablation_example}
traces both agents turn by turn on a task that
\reflact{}+Plan fails and \method{} solves.

\paragraph{What changes inside an episode: fewer hallucinated actions.}
To analyze the temporal effects of the anchoring instruction, we use the
\textbf{invalid-action rate}, the fraction of executed actions the environment
refuses to interpret or apply, as a proxy for \emph{hallucinated actions}. We
pool observations from all episodes regardless of success, and
Appendix~\ref{app:extra_analyses} shows that the pattern reflects grounding
rather than formatting. We report this rate overall and across three temporal
thirds of each episode.
Table~\ref{tab:grounding} gives the values for all six agents on ScienceWorld.

First, grounding errors in reflection-only backbones
(\react{} and \reflact{}) compound over time. For \reflact{} on
Llama-3.1-8B-Instruct, the invalid-action rate rises sharply from 14.5\% to
44.1\% to 71.4\% (Figure~\ref{fig:compounding}, Appendix~\ref{app:moved_tables}), so ungrounded actions become
more frequent, not just more numerous.

Second, \method{} substantially mitigates this error accumulation. On
Llama-3.1-8B-Instruct, \method{} reduces the overall invalid-action rate from
43.3\% (\reflact{}) to 29.9\% and flattens the late-stage surge (from 71.4\% to
49.7\%); \beammethod{} also stays under \reflact{} (34.2\% overall), though above
\method{}. The improvement over \reflact{} holds on GPT-4o as well, both overall
(27.7\% versus 34.2\%) and early (11.2\% versus 19.0\%). On GPT-4o-mini the
reduction is confined to the early third (17.7\% versus 23.3\%): the overall rate
is unchanged (32.6\% versus 32.7\%) and the late third is worse (45.5\% versus
39.0\%).
Appendix~\ref{app:grounding_example} traces one such cascade step by step.

\paragraph{A low invalid-action rate is necessary but not sufficient.}
While a low invalid-action rate might suggest an agent is merely attempting less
demanding actions, Table~\ref{tab:grounding} shows that this metric is only
meaningful when paired with high success rates. On GPT-4o and GPT-4o-mini,
agents without per-turn reflection (e.g., NoThinking and Plan-and-Act) exhibit
the lowest invalid-action rates (20.9\% and 23.3\% on GPT-4o) but also the
lowest success rates (53.6\% and 54.0\%, against 64.0\% for \method{}). Those
low rates are not a matter of acting less: on GPT-4o, NoThinking averages
$24.7$ actions per episode and Plan-and-Act $28.3$, against \method{}'s $20.8$
(Appendix~\ref{app:actioncount}). The difference lies in action quality, not
quantity. In contrast, \method{} and
\beammethod{} reduce invalid
actions and increase success rates at the same time, offering the more
informative within-family comparison.
This is most evident on the constrained Llama-3.1-8B-Instruct, where our
variants achieve both the lowest invalid-action rates and the highest success
rates.

\paragraph{Where the gains show up: procedurally ordered tasks.}
The benefits of improved grounding are non-uniform across tasks.
Figure~\ref{fig:tasktype} (Appendix~\ref{app:moved_tables}) decomposes ScienceWorld into ten task buckets on
Llama-3.1-8B-Instruct (Table~\ref{tab:tasktype11} gives the per-row values),
revealing that gains are most pronounced in tasks requiring strict prerequisite
ordering. For instance, \reflact{} achieves only 0\% on Boil and 5.6\% on
Paint/Mix---the latter of which underperforms even the baseline \react{}
(22.2\%). In contrast, by anchoring reflections to task-relevant sub-goals,
\method{} achieves 55.6\% and 61.1\% on these same tasks, whereas end-goal
reflection can lead to misguided trajectories. The performance margin narrows for
Find/classification (77.5\% for \method{} vs.\ 70.0\% for \reflact{}), where
sub-goal structuring offers minimal advantage for one-step decisions. However,
\reflact{} outperforms \method{} on knowledge-retrieval Lifespan tasks
(66.7\% vs.\ 53.3\%), suggesting that \method{} tends to over-decompose tasks:
these tasks require only two ordered actions (Appendix~\ref{app:nchoice}), yet
the \method{} planner still emits four. For long-horizon exploration (Grow),
no agent exceeds 5.0\%, indicating that the step budget, rather than the
planning mechanism, is the primary bottleneck.
\subsection{Plan Content as a Fundamental Determinant of Backbone Performance (C2)}
\label{sec:plan_quality}

\paragraph{The Primacy of Plan Content.}
Since the two share a backbone (\S3.2), Table~\ref{tab:main} isolates the
planner's impact. Ranked by success rate, \method{} is ahead of \beammethod{} in all
nine (model, benchmark) cells; against \reflact{}, \method{} is ahead in all
nine, whereas \beammethod{} is ahead in five, which suggests that the planner
limits the backbone's potential: if candidate plans miss necessary steps,
increasing the search budget will not help. We turn to plan content for the
rest of this section (Table~\ref{tab:procedure}).

\paragraph{Structural Deficiencies in Planning.}
Qualitative analysis attributes these failures to a structural deficiency:
deficient plans often merely restate the goal rather than prescribing physical
transformations. A ``Melt'' plan may include a sub-goal such as
``change its state to liquid,'' restating the task description instead of
naming the required operation (applying a heat source). Corroborated by our
observations on Llama-3.1-8B-Instruct, where the agent earns partial reward on
14 of the 18 Freeze and Melt episodes yet solves none, the failure lies in the
plan's content rather than execution errors. Because these critical steps are
entirely absent, search-based reordering cannot resolve the failure.

\paragraph{Procedural Guidance as a Remedy.}
To address this, we investigated whether explicit procedural prompting could
recover these missing steps. This procedural prompting is applied
\emph{only to the Planner}, never to the agent prompt, so the comparison stays
fair. As shown in Table~\ref{tab:procedure} (Appendix~\ref{app:moved_tables}),
introducing it for state-change tasks (Appendix~\ref{app:sci_planner}) yielded
clear improvements, increasing success on the targeted state-change episodes by
up to $16.7$ points on GPT-4o.
Because the agent never sees the procedure, these improvements are driven by the
Planner through the sub-goals it generates. Llama-3.1-8B-Instruct's single-path
planner is the exception, losing $1.4$ points overall and $5.6$ on state-change,
because the small LLM cannot turn the added rule into a correctly ordered
sub-goal plan; the Beam Search Planner instead achieves this on the targeted
episodes, matching or surpassing single-path performance across all tested
models. The bottleneck therefore lies in the absence of the required
operations, not the search mechanism: explicit prompting supplies the
operation, though the search mechanism remains necessary for the weakest
model to use it.

\paragraph{Limitations of Search-Based Optimization.}
Without the explicit rule, \beammethod{} does not outperform the single-path
baseline, which we attribute not to the search mechanism itself but to how
candidates are scored. Our analysis points to two issues. First, the
self-evaluation judge (Equation~\ref{eq:selfeval}) lacks environmental
grounding, relying exclusively on the LLM's internal knowledge. Second, the
scorer is backward-looking: its rubric (Appendix~\ref{app:beam}) prioritizes
the local consistency of existing sub-goals over future task completion, which
is assessed only at the final depth. Locally well-formed prefixes can therefore
score highly even without essential future operations, so high-scoring but
flawed prefixes are preserved and increasing the beam width merely reorders
existing candidates rather than introducing the missing operations required for
success. This bottleneck is most evident in Jericho, where the gap between
plausible-sounding sub-goals and executable actions is widest, not from the
beam width itself: a budget-comparable greedy search (up to
48 calls, against beam's 42) lands at $27.0$\% vs.\ \beammethod{}'s $26.5$\% on
Llama-3.1-8B-Instruct (Table~\ref{tab:greedy}), though we tested no other
configuration on the larger models. While search may occasionally recover
isolated operations (e.g., \texttt{melt caesium}; Appendix~\ref{app:plan_examples}),
it cannot enhance aggregate performance independently.

  \section{Conclusion}
  \label{sec:conclusion}

  In this paper, we presented \method{}, a reasoning backbone that decomposes
  task goals into a sub-goal plan and anchors each reflection
  to the active sub-goal. To investigate the impact of planning quality, we
  also introduced \beammethod{}, which uses a self-evaluation-guided beam
  search. Across ALFWorld, ScienceWorld, and Jericho, \method{}
  outperformed \react{}, Plan-and-Act, and \reflact{} in nearly all settings,
  with the largest
  gains on the smallest model (e.g., $+14.9$\,pp on ALFWorld and
  $+8.0$\,pp on ScienceWorld using Llama-3.1-8B-Instruct).

  Our analysis reveals two primary drivers of this improvement. First,
  anchoring reflections to sub-goals substantially enhances agent grounding,
  reducing the invalid-action rate and improving performance especially on
  procedurally ordered tasks. Ablation studies confirm that this benefit stems from the
  anchoring instruction itself rather than the mere presence of a sub-goal plan.
  Second, we suggest that plan content is a fundamental determinant of
  backbone performance. Search-based optimization only
  yields benefits when candidate plans explicitly include the required
  operations. Consequently, improving the descriptive quality of sub-goals is a
  necessary prerequisite for effective plan search, rather than a substitute
  for it.

  These findings suggest a dual design principle for long-horizon agents: (i)
  strengthening reasoning backbones requires refining their reflection targets,
  and (ii) optimizing planners necessitates prioritizing sub-goal content to
  enable effective search.

  \paragraph{Limitations and Future Work.}
  Several limitations remain. First, \method{} fixes the sub-goal plan at
  $t=0$, meaning it cannot dynamically generate new sub-goals mid-episode;
  implementing a re-planning trigger is a natural next step. Second, some of
  our margins rest on few episodes: a single game is worth 5.0 points on the
  20-game Jericho suite, and a single episode moves rates from splits as
  small as nine by $11.1$ points. We also fix the plan
  length at $N=4$ for every task (Section~\ref{sec:task_analysis},
  Appendix~\ref{app:nchoice}). Finally, our modular planner can be
  replaced by environment-aware models, though it incurs up to 42 LLM calls
  per episode.


\section*{Reproducibility Statement}
All agents, environments, and evaluation code are re-implemented in a single
codebase and run under identical conditions, described in
Section~\ref{sec:experiments}. Every run shares the same base LLM per group, the
same stop-word set, the same flat-text in-context learning format, and a single
fixed random seed (42). Appendices~\ref{app:setup} and~\ref{app:maxsteps} give
the exact settings, the corrected success criteria, and the per-task step
budgets.
Appendix~\ref{app:beam} gives the full beam-search planning algorithm, its
hyperparameters ($N=4$, $K=2$, $S=3$), the self-evaluation scoring values, and
the self-evaluation prompt. Every system prompt, in-context example, and planner
prompt is listed in Appendices~\ref{app:prompts},~\ref{app:icl},
\ref{app:planner_prompts},~\ref{app:sci_planner}, and~\ref{app:jericho_planner}.
The per-group results behind every aggregate in Table~\ref{tab:main}, and the
dispersions it reports, are given in Appendices~\ref{app:alfworld_cat}
and~\ref{app:sciworld_cat}.

\section*{Ethics Statement}
This work evaluates LLM agents in three simulated text environments (ALFWorld,
ScienceWorld, and Jericho). It involves no human subjects, no personal or
sensitive data, and no deployment in a physical or safety-critical setting. The
method changes how an agent reasons; it does not grant the agent new capabilities
or tool access. We use publicly released benchmarks under their original
licenses, together with commercially available or openly released models. We see
no direct path from this work to harmful application beyond the general risks
already associated with more capable LLM agents.

\section*{AI Use Statement}
We used LLM-based assistants during this project, and disclose their role below.
We did not use them to generate synthetic data, to develop theoretical results or
proofs, or to implement the agent methods and evaluation harness; the remaining
disclosure categories in the ICLR AI policy do not apply to this work. We have
reviewed all AI-assisted work. The authors take full responsibility for all
content, including any text, figures, or analysis that an assistant helped
produce.

\paragraph{Research methodology and analysis.}
An assistant was used to discuss and stress-test the framing of the analyses in
Section~\ref{sec:discussion}. It proposed robustness checks---such as stratifying the
invalid-action rate by episode outcome and aggregating the four state-change
task types---wrote the scripts that computed them, and helped interpret the
resulting numbers. The authors reviewed those scripts and checked their output
against the run logs. All agent runs were designed, executed, and verified by
the authors, and every number reported in this paper was computed from our own
experiment logs.

\paragraph{Literature retrieval and citation checking.}
An assistant was used to retrieve and summarize the abstracts of the works we
cite and to flag mismatches between a citation and the claim it supported. Every
flagged case was checked by the authors against the source, and any replacement
citation was verified against that source before use.

\paragraph{Writing.}
An assistant was used to shorten and simplify prose, to split long sentences, to
align terminology, and to check internal consistency between the text, the
tables, and the figures. It did not generate any claim, result, or citation that
the authors did not verify against the underlying data or the cited source.

\paragraph{Code.}
An assistant wrote most of the analysis and plotting scripts, including the one
that computes the ordered-chain length behind our choice of $N=4$
(Appendix~\ref{app:nchoice}); the authors reviewed them and verified their
output. The agent implementations and the evaluation harness were written and
checked by the authors.

\bibliography{references}
\bibliographystyle{iclr2027_conference}

\appendix

\section{Framework Comparison}
\label{app:comparison_sec}

Table~\ref{tab:comparison} places \method{} and \beammethod{} among the planning
and reflection frameworks reviewed in Section~\ref{sec:related}. It separates the
two components that prior work supplies on their own---an upfront plan and a
per-turn reflection---from the coupling that makes the plan the target of that
reflection, which is the axis on which our two agents differ from the rest.

\begin{table}[h]
\centering
\caption{
  We compare agent frameworks along three design axes: \textbf{Plan}, whether
  the agent performs structured upfront goal decomposition; \textbf{Turn-Reflect},
  whether it performs per-turn state--goal reflection; and \textbf{Plan+Reflect},
  whether the two components are tightly coupled so the plan anchors the
  reflection.
  MPO provides a meta plan as context but does not modify the reflection
  target, so it has Plan but not Turn-Reflect or Plan+Reflect.
}
\label{tab:comparison}
\small
\setlength{\tabcolsep}{4pt}
\begin{tabular}{lccc}
\toprule
Method & Plan & Turn-Reflect & Plan+Reflect \\
\midrule
\react{} \citep{yao2023react}          & \texttimes    & \texttimes    & \texttimes \\
Reflexion \citep{shinn2023reflexion}   & \texttimes    & \texttimes    & \texttimes \\
WKM \citep{qiao2024wkm}                & \texttimes    & \texttimes    & \texttimes \\
Plan-and-Act \citep{erdogan2025planact}& \checkmark    & \texttimes    & \texttimes \\
MPO \citep{mpo2025}                    & \checkmark    & \texttimes    & \texttimes \\
LATS \citep{zhou2024lats}              & Tree/MCTS     & \texttimes    & \texttimes \\
\reflact{} \citep{reflact2025}         & \texttimes    & \checkmark    & \texttimes \\
ToT / GoT \citep{yao2023tot, besta2024got} & Tree/Graph & \texttimes  & \texttimes \\
SelfEval-Beam \citep{xie2023selfevalguided} & Beam     & \texttimes   & \texttimes \\
\midrule
\method{} (ours)     & Single-pass   & Sub-goal & \checkmark \\
\beammethod{} (ours) & Beam+SelfEval & Sub-goal & \checkmark \\
\bottomrule
\end{tabular}
\end{table}

\section{Choosing the Plan Length $N$}
\label{app:nchoice}

For each of ScienceWorld's 24 task types, we take the environment's own
gold action sequence (\texttt{generateGoldPath=True}) and compute the length
of the longest chain of actions that the sequence forces into a fixed order:
the sub-sequence of distinct actions that cannot be reordered without
breaking a prerequisite. This length is the natural analogue of the number of
ordered milestones a sub-goal plan should name, since our sub-goals are
likewise ordered to respect the environment's prerequisites
(Section~\ref{sec:method:framework}). Table~\ref{tab:nchoice} gives this
length for every type. Weighted by the number of test episodes each type
contributes, the mean is $3.79$ and the median and mode are both $4$, which is
why we set $N=4$. The values range from $2$ (the Lifespan and Life-Stage tasks,
which only require locating and comparing organisms) to $8$ (Chemistry Mix,
which chains several sequential mixing steps), so $N=4$ is a compromise tuned
to the center of this distribution rather than a value that fits every type
exactly.

\begin{table}[h]
\centering
\caption{Length of the longest forced-order action chain in the ScienceWorld
  gold solution, by task type, alongside the number of test episodes each
  type contributes.}
\label{tab:nchoice}
\small
\begin{tabular}{lcc}
\toprule
Task type & $n$ & Chain length \\
\midrule
Boil                              &  9 & 4 \\
Change state of matter            &  9 & 4 \\
Freeze                             &  9 & 4 \\
Melt                                & 9 & 5 \\
Measure melting point              & 10 & 4 \\
Use thermometer                    & 10 & 3 \\
Power component                    &  5 & 3 \\
Power (renewable vs.\ non-renewable) & 5 & 5 \\
Test conductivity                  & 10 & 6 \\
Test conductivity (unknown)        & 10 & 6 \\
Find animal                        & 10 & 4 \\
Find living thing                  & 10 & 4 \\
Find non-living thing              & 10 & 4 \\
Find plant                         & 10 & 4 \\
Grow fruit                         & 10 & 3 \\
Grow plant                         & 10 & 3 \\
Chemistry mix                      &  8 & 8 \\
Mix paint (secondary)              &  9 & 4 \\
Mix paint (tertiary)               &  9 & 2 \\
Lifespan (longest-lived)           & 10 & 2 \\
Lifespan (longest then shortest)   & 10 & 2 \\
Lifespan (shortest-lived)          & 10 & 2 \\
Identify life stages 1             &  5 & 2 \\
Identify life stages 2             &  4 & 2 \\
\bottomrule
\end{tabular}
\end{table}

\section{BeamSGG-ReflAct Planning Details}
\label{app:beam}

Algorithm~\ref{alg:beam} gives the full beam-search planning procedure of
\beammethod{} (Section~\ref{sec:method:beam}).
The self-evaluation score used in the multiplicative plan score
(Equation~\ref{eq:beam_score}) asks the judge to rate the partial plan on a
continuous scale in $[0,1]$ against a fixed rubric:
\begin{equation}
  \textsc{SelfEval}(\boldsymbol{\rho}) =
  \left\{\begin{array}{ll}
    0.01 & \text{duplicate} \\
    \mathrm{clip}_{[0.05,\,1.0]}\!\left(v\right) & \text{parsed rating } v \\
    0.30 & \text{no rating parsed}
  \end{array}\right.
  \label{eq:selfeval}
\end{equation}
A candidate scores $0.01$ when $g_d$ duplicates any sub-goal in
$\boldsymbol{\rho}_{1:d-1}$, which is detected without an LLM call. A rating given on a
$0$--$10$ or $0$--$100$ scale is rescaled, and the result is clipped to
$[0.05, 1.0]$ so that no single weak step can zero out an entire plan. We ask for
a continuous rating rather than a binary verdict because a binary judge collapses
every acceptable candidate onto one value and so cannot rank candidates against
one another.
The rubric asks the LLM to check several criteria:
(1) the sub-goal describes \emph{what} to achieve and follows any ordering
instructions in $u$;
(2) no prerequisite step is skipped;
(3) a ``focus on'' sub-goal appears only after the target object has been
obtained;
(4) the final sub-goal achieves the stated objective; and
(5) variable elements are described generically.
The closing instruction that requests the rating is reproduced verbatim below,
and the full per-environment prompts are in
Appendices~\ref{app:planner_prompts},~\ref{app:sci_planner},
and~\ref{app:jericho_planner}.

\begin{promptbox}{Self-evaluation: closing instruction}
\small
\begin{lstlisting}
Rate how well the Partial plan satisfies the criteria above on a continuous scale from 0.0 to 1.0 (1.0 = fully correct and well-ordered, 0.0 = clearly wrong). Use the full range: a plan that satisfies most but not all criteria should receive an intermediate value.
Answer with a single decimal number only (e.g., 0.82). No explanation.
\end{lstlisting}
\end{promptbox}

\paragraph{How the $S$ candidates differ.}
Decoding is deterministic, so a beam's $S$ candidates are drawn one at a time
rather than in parallel, and each request names what has already been produced.
The expansion prompt is followed by the in-context example, the task, the
sub-goals fixed so far, up to two avoidance lines, and a closing request for
step $d$ of $N$:

\begin{promptbox}{Avoidance lines in the expansion prompt}
\small
\begin{lstlisting}
Do not repeat any sub-goal already in the current plan: {sub-goals fixed so far}
Do not repeat any of the following proposed candidates: {candidates already drawn at this depth}
\end{lstlisting}
\end{promptbox}

The $i$-th request therefore lists the first $i-1$ candidates, which is what
makes them differ under deterministic decoding. Self-evaluation is independent
across candidates and runs in parallel. Before the top-$K$ cut, candidates that
reached the same sub-goal sequence are collapsed to the highest-scoring one.

\begin{algorithm}[h]
\caption{\beammethod{} Planning}
\label{alg:beam}
\begin{algorithmic}[1]
  \REQUIRE Task instruction $u$, beam width $K$, samples per beam $S$, depth $N$
  \STATE $\mathcal{B}_0 \leftarrow \{()\}$
  \FOR{$d = 1$ \TO $N$}
    \STATE $\mathcal{C} \leftarrow \emptyset$
    \FOR{each partial plan $\boldsymbol{\rho} \in \mathcal{B}_{d-1}$}
      \STATE avoid\_set $\leftarrow \emptyset$
      \FOR{$i = 1$ \TO $S$}
        \STATE $g_d^{(i)} \leftarrow \mathcal{M}(\text{expand\_prompt},\; u,\; \boldsymbol{\rho},\; \text{avoid\_set})$
        \STATE avoid\_set $\leftarrow$ avoid\_set $\cup\, \{g_d^{(i)}\}$
        \STATE $\mathcal{C} \leftarrow \mathcal{C} \cup \{(\boldsymbol{\rho},\, g_d^{(i)})\}$
      \ENDFOR
    \ENDFOR
    \STATE \textit{// Score all candidates in parallel via self-evaluation}
    \FOR{each $\boldsymbol{\rho}' \in \mathcal{C}$}
      \STATE $\mathrm{score}(\boldsymbol{\rho}') \leftarrow \mathrm{score}(\boldsymbol{\rho}'_{1:d-1}) \times \textsc{SelfEval}(\boldsymbol{\rho}')$
    \ENDFOR
    \STATE $\mathcal{B}_d \leftarrow \mathrm{top}\text{-}K(\mathcal{C},\; \mathrm{score})$
  \ENDFOR
  \RETURN $\displaystyle\arg\max_{\boldsymbol{\rho}\,\in\,\mathcal{B}_N} \mathrm{score}(\boldsymbol{\rho})$
\end{algorithmic}
\end{algorithm}

\section{Step Budgets and Episode Termination}
\label{app:maxsteps}

An episode ends when the agent solves the task, when it reaches the step budget,
or---on ScienceWorld---when it emits ten consecutive actions the environment
cannot parse. We cap ALFWorld and Jericho at \textbf{30} steps per episode,
following AgentBoard \citep{ma2024agentboard} for Jericho. ScienceWorld uses a
per-task budget, listed in Table~\ref{tab:maxsteps}, because its task types
differ widely in how many actions a gold solution needs.

ScienceWorld additionally terminates an episode after \textbf{ten consecutive}
observations of \texttt{No known action matches that input}; a single valid
action resets the counter. This rule caps how long an agent can loop on a
phrasing the parser rejects, and it bounds the late-episode invalid-action rates
reported in Section~\ref{sec:task_analysis}.

Responses missing the \texttt{Action:} tag trigger an \texttt{Error Input}
observation rather than an environment execution, and this is uniform across all
three benchmarks. Such turns consume step budget without advancing the
environment state. On ScienceWorld they do not count toward the ten consecutive
unparsable actions above, which are the environment's own refusals.

\begin{table}[h]
\centering
\caption{
  Per-task step budgets for ScienceWorld's 24 task types. An episode also ends
  early on success or after ten consecutive unparsable actions.
}
\label{tab:maxsteps}
\small
\setlength{\tabcolsep}{6pt}
\begin{tabular}{lr}
\toprule
Task Type & Max Steps \\
\midrule
\texttt{task-10-measure-melting-point-(known-substance)} & 120 \\
\texttt{task-1-boil} & 100 \\
\texttt{task-1-change-the-state-of-matter-of} & 80 \\
\texttt{task-1-freeze} & 80 \\
\texttt{task-1-melt} & 80 \\
\texttt{task-4-grow-fruit} & 60 \\
\texttt{task-5-chemistry-mix} & 60 \\
\texttt{task-10-use-thermometer} & 30 \\
\texttt{task-2-power-component-(renewable-vs-nonrenewable-energy)} & 30 \\
\texttt{task-2a-test-conductivity} & 30 \\
\texttt{task-2a-test-conductivity-of-unknown-substances} & 30 \\
\texttt{task-4-grow-plant} & 30 \\
\texttt{task-5-chemistry-mix-paint-(tertiary-color)} & 30 \\
\texttt{task-7-identify-life-stages-1} & 30 \\
\texttt{task-7-identify-life-stages-2} & 30 \\
\texttt{task-2-power-component} & 20 \\
\texttt{task-3-find-animal} & 15 \\
\texttt{task-3-find-living-thing} & 15 \\
\texttt{task-3-find-non-living-thing} & 15 \\
\texttt{task-3-find-plant} & 15 \\
\texttt{task-5-chemistry-mix-paint-(secondary-color)} & 15 \\
\texttt{task-6-lifespan-(longest-lived-then-shortest-lived)} & 12 \\
\texttt{task-6-lifespan-(longest-lived)} & 10 \\
\texttt{task-6-lifespan-(shortest-lived)} & 10 \\
\bottomrule
\end{tabular}
\end{table}

\section{Experimental Settings}
\label{app:setup}

This appendix records the settings summarized in Section~\ref{sec:experiments}.
All models decode with \texttt{temperature=0} and \texttt{max\_tokens=512}, and
no method updates any model weights, so agents differ from one another only in
their prompts and control flow. To ensure consistency, all agents share the
same base LLM, stop-word set, and in-context format within a single codebase.
Environments and test splits follow \citet{reflact2025} and \citet{mpo2025} for
ALFWorld and ScienceWorld and \citet{ma2024agentboard} for Jericho, re-implemented
in our own codebase rather than reusing the original code. To mitigate known
success-flag bugs in ScienceWorld and Jericho \citep{reflact2025} (ScienceWorld's
\texttt{getCompleted()} can return \texttt{True} at low reward values, and Jericho
shows the same false positive after the agent has died), we define success via
progress rewards: at least $0.7$ for ScienceWorld and exactly $1.0$ for Jericho;
ALFWorld uses its native binary flag. Appendix~\ref{app:maxsteps} gives the
per-task step budgets, and \beammethod{} is the one agent that issues extra model
calls beyond the per-turn policy, at most 42 planner calls per episode
(Appendix~\ref{app:beam}).

We exclude Reflexion \citep{shinn2023reflexion} due to its multi-episode reset
mechanism, which is incompatible with our single-episode online setting. MPO and
LATS are discussed in Section~\ref{sec:related} rather than run:
\citet{reflact2025} already measure MPO in this setting and report that adding
meta-plans to a reflection backbone barely changes performance, with a small drop
on Llama-3.1-8B-Instruct; LATS is excluded because its environment-rollout search
exceeds our budget of one call per episode with no environment interaction.

\section{ALFWorld Success Rate by Task Type}
\label{app:alfworld_cat}

Table~\ref{tab:alfcat} breaks the ALFWorld results of Table~\ref{tab:main} down by
the benchmark's six standard task types, which we recover from the goal text of
each episode.
The \emph{SD} column is the standard deviation of the six rates and is the
dispersion reported alongside the ALFWorld column of Table~\ref{tab:main}.
Pick Two \& Place is the hardest type for every model, and it is where the
weakest backbones collapse.

\begin{table}[h]
\centering
\caption{
  We report ALFWorld success rate (\%) by task type, for all three models.
  Columns abbreviate the six standard types: \emph{Pick} is Pick \& Place
  ($n{=}24$), \emph{Look} is Look at Object in Light ($n{=}18$), \emph{Clean}
  is Clean \& Place ($n{=}31$), \emph{Heat} is Heat \& Place ($n{=}23$),
  \emph{Cool} is Cool \& Place ($n{=}21$), and \emph{Two} is Pick Two \& Place
  ($n{=}17$).
  \emph{All} is the overall rate over the 134 episodes and matches
  Table~\ref{tab:main}, and \emph{SD} is the standard deviation across the six
  types. We bold the best result per column within each model group.
}
\label{tab:alfcat}
\scriptsize
\setlength{\tabcolsep}{3pt}
\begin{tabular}{llcccccccc}
\toprule
Model & Agent & Pick & Look & Clean & Heat & Cool & Two & All & SD \\
\midrule
\multirow{6}{*}{GPT-4o}
& NoThinking       & 87.5 & \textbf{100.0} & 83.9 & 52.2 & 85.7 & 58.8 & 78.4 & 16.8 \\
& \react{}         & 91.7 & 94.4 & 80.6 & 73.9 & 81.0 & 58.8 & 80.6 & 11.8 \\
& Plan-and-Act     & 87.5 & 94.4 & 77.4 & 65.2 & 90.5 & \textbf{76.5} & 81.3 & 9.9 \\
& \reflact{}       & \textbf{95.8} & \textbf{100.0} & \textbf{90.3} & 78.3 & 95.2 & \textbf{76.5} & 89.6 & 8.9 \\
& \beammethod{}    & \textbf{95.8} & 88.9 & 87.1 & 78.3 & 95.2 & 52.9 & 84.3 & 14.7 \\
& \method{}        & \textbf{95.8} & \textbf{100.0} & 87.1 & \textbf{82.6} & \textbf{100.0} & \textbf{76.5} & 90.3 & 8.9 \\
\midrule
\multirow{6}{*}{GPT-4o-mini}
& NoThinking       & 83.3 & 61.1 & 29.0 & 52.2 & 19.0 & 17.6 & 44.0 & 24.0 \\
& \react{}         & 75.0 & 61.1 & 38.7 & 34.8 & 38.1 & 23.5 & 45.5 & 17.4 \\
& Plan-and-Act     & 50.0 & 50.0 & 54.8 & 34.8 & 33.3 & \textbf{29.4} & 43.3 & 9.8 \\
& \reflact{}       & 79.2 & 66.7 & 58.1 & \textbf{82.6} & 61.9 & 17.6 & 62.7 & 21.3 \\
& \beammethod{}    & 79.2 & 66.7 & \textbf{83.9} & 69.6 & \textbf{90.5} & 11.8 & 70.1 & 26.0 \\
& \method{}        & \textbf{87.5} & \textbf{77.8} & 77.4 & 73.9 & 81.0 & 23.5 & 72.4 & 21.3 \\
\midrule
\multirow{6}{*}{\shortstack[l]{Llama-3.1\\-8B}}
& NoThinking       & 58.3 & 66.7 & 19.4 & 21.7 & 0.0 & 5.9 & 28.4 & 25.2 \\
& \react{}         & 33.3 & 27.8 & \textbf{32.3} & 26.1 & 33.3 & 23.5 & 29.9 & 3.8 \\
& Plan-and-Act     & 50.0 & 38.9 & 29.0 & 8.7 & 33.3 & 17.6 & 29.9 & 13.5 \\
& \reflact{}       & 66.7 & \textbf{77.8} & 19.4 & 4.3 & 14.3 & \textbf{47.1} & 35.8 & 27.5 \\
& \beammethod{}    & \textbf{75.0} & 72.2 & 16.1 & \textbf{43.5} & 57.1 & 23.5 & 46.3 & 22.5 \\
& \method{}        & 62.5 & 61.1 & 29.0 & 39.1 & \textbf{85.7} & 35.3 & 50.7 & 19.6 \\
\bottomrule
\end{tabular}
\end{table}

\section{ScienceWorld Success Rate by Task Category}
\label{app:sciworld_cat}

Table~\ref{tab:scicat} breaks the ScienceWorld results of Table~\ref{tab:main}
down by the nine categories of the ScienceWorld split, which group its 24 task
types by task identifier (\texttt{task-1} through \texttt{task-10}), for all
three models. The \emph{All} column and the \emph{SD} column are the two numbers
Table~\ref{tab:main} reports for ScienceWorld: \emph{SD} is the dispersion
printed after each success rate there.
Two patterns recur across models. The state-change category, which pools the
Boil, Freeze, Melt, and Change-State task types, is where the spread between
methods is widest, and Grow stays near zero for every agent and model,
confirming that its bottleneck is the step budget rather than the reasoning
backbone.

\begin{table}[h]
\centering
\caption{
  We report ScienceWorld success rate (\%) by task category, for all three
  models. The nine columns group the 24 ScienceWorld task types by task
  identifier: \emph{St-Chg} ($n{=}36$), \emph{Elec} (10), \emph{Cond} (20),
  \emph{Find} (40),
  \emph{Grow} (20), \emph{Chem} (26), \emph{Lifesp} (30), \emph{LifeSt} (9),
  and \emph{Meas} (20).
  \emph{All} is the overall rate over the 211 episodes and \emph{SD} the
  standard deviation across the nine categories; both match the corresponding
  ScienceWorld entry of Table~\ref{tab:main}. We bold the best result per
  category column within each model group, with ties bolded both, and leave
  columns on which every agent ties unmarked.
}
\label{tab:scicat}
\scriptsize
\setlength{\tabcolsep}{2.5pt}
\begin{tabular}{llccccccccccc}
\toprule
Model & Agent & St-Chg & Elec & Cond & Find & Grow & Chem & Lifesp & LifeSt & Meas & All & SD \\
\midrule
\multirow{6}{*}{GPT-4o}
& NoThinking       & 47.2 & \textbf{10.0} & 15.0 & 82.5 & 0.0 & 73.1 & 80.0 & \textbf{44.4} & 60.0 & 53.6 & 29.4 \\
& \react{}         & 55.6 & 0.0 & 10.0 & 97.5 & 0.0 & 76.9 & \textbf{86.7} & \textbf{44.4} & 70.0 & 59.2 & 35.6 \\
& Plan-and-Act     & 41.7 & 0.0 & 5.0 & 97.5 & 5.0 & 65.4 & 80.0 & \textbf{44.4} & 65.0 & 54.0 & 33.4 \\
& \reflact{}       & 61.1 & 0.0 & 15.0 & 87.5 & 0.0 & 46.2 & \textbf{86.7} & \textbf{44.4} & 75.0 & 55.5 & 32.7 \\
\ourrow & \beammethod{}    & \textbf{66.7} & \textbf{10.0} & \textbf{50.0} & 87.5 & 0.0 & 73.1 & 80.0 & 33.3 & \textbf{90.0} & 63.5 & 31.5 \\
\ourrow & \method{}        & 55.6 & 0.0 & \textbf{50.0} & \textbf{100.0} & \textbf{10.0} & \textbf{84.6} & 76.7 & 0.0 & \textbf{90.0} & 64.0 & 37.4 \\
\midrule
\multirow{6}{*}{GPT-4o-mini}
& NoThinking       & 5.6 & 0.0 & 0.0 & 55.0 & 0.0 & 3.8 & 60.0 & 11.1 & 0.0 & 20.9 & 23.0 \\
& \react{}         & 25.0 & 0.0 & \textbf{15.0} & \textbf{100.0} & 0.0 & 30.8 & 73.3 & 0.0 & 20.0 & 40.8 & 33.0 \\
& Plan-and-Act     & 16.7 & 0.0 & 5.0 & 97.5 & 0.0 & 7.7 & \textbf{76.7} & \textbf{22.2} & 15.0 & 36.0 & 33.4 \\
& \reflact{}       & 41.7 & 0.0 & 10.0 & \textbf{100.0} & 0.0 & 3.8 & 73.3 & 0.0 & 25.0 & 40.3 & 34.5 \\
\ourrow & \beammethod{}    & 25.0 & 0.0 & 10.0 & 85.0 & 0.0 & 61.5 & 73.3 & 0.0 & \textbf{45.0} & 43.6 & 31.9 \\
\ourrow & \method{}        & \textbf{52.8} & 0.0 & 0.0 & 95.0 & 0.0 & \textbf{65.4} & 60.0 & 11.1 & 10.0 & 45.0 & 33.8 \\
\midrule
\multirow{6}{*}{\shortstack[l]{Llama-3.1\\-8B}}
& NoThinking       & 2.8 & 0.0 & 10.0 & 55.0 & 0.0 & 0.0 & 3.3 & 0.0 & 10.0 & 13.3 & 16.7 \\
& \react{}         & 2.8 & 0.0 & 15.0 & 45.0 & \textbf{5.0} & 15.4 & 40.0 & 0.0 & 0.0 & 18.5 & 16.4 \\
& Plan-and-Act     & 0.0 & 0.0 & \textbf{20.0} & 37.5 & \textbf{5.0} & 7.7 & 3.3 & 0.0 & 10.0 & 11.8 & 11.7 \\
& \reflact{}       & 11.1 & 0.0 & 0.0 & 70.0 & \textbf{5.0} & 3.8 & \textbf{66.7} & 0.0 & 0.0 & 25.6 & 27.5 \\
\ourrow & \beammethod{}    & 16.7 & 0.0 & 15.0 & 70.0 & 0.0 & 3.8 & 53.3 & 0.0 & 10.0 & 26.5 & 24.0 \\
\ourrow & \method{}        & \textbf{22.2} & 0.0 & 0.0 & \textbf{77.5} & \textbf{5.0} & \textbf{42.3} & 53.3 & 0.0 & \textbf{20.0} & 33.6 & 26.2 \\
\bottomrule
\end{tabular}
\end{table}

Figure~\ref{fig:tasktype} regroups the same episodes into ten reporting
buckets rather than these nine categories, so that the state-change category
appears as its four task types and the paint-mixing episodes appear on their
own; the remaining categories are pooled into \emph{Other}.
Table~\ref{tab:tasktype11} gives the finer split the figure is built from, for
the same model and the same four backbones.
Reading it beside the figure makes two things visible that the pooled
\emph{Other} bar hides.
First, \beammethod{}'s nonzero rate on \emph{Other} comes from four episodes
out of forty-seven: three electrical or conductivity episodes and one
non-paint chemistry episode.
Second, no agent solves a single life-stage episode, so that category does not
separate the two planners at all.
On the state-change tasks, \beammethod{} solves one of nine Melt episodes and
\method{} none; these are the tasks where Section~\ref{sec:plan_quality} shows
that stating the required procedure in the plan, rather than the planner that
produced it, drives the difference.

\begin{table}[h]
\centering
\caption{
  We report ScienceWorld success rates (\%) with Llama-3.1-8B-Instruct at the
  granularity Figure~\ref{fig:tasktype} is drawn from: the state-change
  category as its four task types, the chemistry category as its paint and
  non-paint episodes, and every other row one of those categories.
  \emph{Electricity} pools the power-component and conductivity categories, as
  the figure does.
  The figure's \emph{Paint\,/\,Mix} bar is the \emph{Paint\,/\,Mix} row here,
  and its \emph{Other} bar pools \emph{Electricity}, \emph{Life Stages}, and
  \emph{Chemistry mix}.
  Totals match Table~\ref{tab:main}; \#\,zero counts the rows an agent never
  solves. We bold the best result per row and leave rows on which no agent
  succeeds unmarked.
}
\label{tab:tasktype11}
\small
\setlength{\tabcolsep}{4pt}
\begin{tabular}{lrcccc}
\toprule
Task Type               & $n$ & \react{} & \reflact{} & \beammethod{} & \method{} \\
\midrule
Boil                    &   9 & 0.0 & 0.0 & 22.2 & \textbf{55.6} \\
Freeze                  &   9 & 0.0 & 0.0 & 0.0 & 0.0 \\
Melt                    &   9 & 0.0 & \textbf{33.3} & 11.1 & 0.0 \\
Change-State            &   9 & 11.1 & 11.1 & \textbf{33.3} & \textbf{33.3} \\
Paint / Mix             &  18 & 22.2 & 5.6 & 0.0 & \textbf{61.1} \\
Chemistry mix           &   8 & 0.0 & 0.0 & \textbf{12.5} & 0.0 \\
Find (classification)   &  40 & 45.0 & 70.0 & 70.0 & \textbf{77.5} \\
Measure                 &  20 & 0.0 & 0.0 & 10.0 & \textbf{20.0} \\
Electricity             &  30 & \textbf{10.0} & 0.0 & \textbf{10.0} & 0.0 \\
Grow                    &  20 & \textbf{5.0} & \textbf{5.0} & 0.0 & \textbf{5.0} \\
Lifespan                &  30 & 40.0 & \textbf{66.7} & 53.3 & 53.3 \\
Life Stages             &   9 & 0.0 & 0.0 & 0.0 & 0.0 \\
\midrule
\textbf{Total}          & 211 & 18.5 & 25.6 & 26.5 & \textbf{33.6} \\
\midrule
SD across rows          &     & 15.5 & 25.1 & 21.8 & 28.0 \\
\#\,zero rows           &     & 6 & 6 & \textbf{4} & 5 \\
\bottomrule
\end{tabular}
\end{table}

\section{Reflection Anchoring: \reflact{}+Plan vs.\ \method{} (Example)}
\label{app:ablation_example}

This example accompanies the ablation of Section~\ref{sec:task_analysis}.
Both agents receive a sub-goal plan for the same ScienceWorld episode
(Llama-3.1-8B-Instruct, \texttt{boil mercury}) from the same single-path
planner, generated independently for each agent, and differ in how the
per-turn reflection is instructed and demonstrated.
\reflact{}+Plan scores 0.05 and \method{} solves the task.
The instructions below are the ScienceWorld ones
(Appendix~\ref{app:prompts}), quoted without the output-format clause they
share. \reflact{}+Plan carries \reflact{}'s released wording unchanged, so the
contrast is between two complete per-turn instructions given plans from the
same planner.

\paragraph{The per-turn instruction.}\mbox{}

\begin{promptbox}{\reflact{}+Plan (end-goal reflection)}
\small\raggedright
\textit{``For each of your turns, you will be given the observation from the
previous turn. You should first reflect on the agent's state, including the
location, inventory, and focused object, \textbf{in relation to the task goal}.
Then, output the action for this turn.\,\ldots''}
\end{promptbox}

\begin{promptbox}{\method{} (sub-goal anchored)}
\small\raggedright
\textit{``For each of your turn, you will be given the observation of the last
turn. \textbf{Referring to the provided sub-goals, determine which sub-goal you
are currently working on based on your progress so far.} Then reflect in one
sentence on your current state and \textbf{progress toward the current
sub-goal}, and then output the action for this turn.\,\ldots''}
\end{promptbox}

\paragraph{Resulting reflections.}
The plans the two agents receive are near-identical in content, both ending in a
heating step.
Their reflections are not.

\begin{promptbox}{\reflact{}+Plan rollout, fails (score 0.05)}
\small
\textit{Plan given:} 1.\ Find the container holding mercury; 2.\ Focus on the
mercury; 3.\ Take actions that will cause the mercury to boil or combust;
4.\ Observe the resulting state of the mercury.
\begin{enumerate}
  \item ``To solve this task, I need to find a container with mercury, focus on
    it, and then take actions to change its state of matter\,\ldots''
    \hfill \emph{(restates the whole plan)}
  \item ``Currently, I am in the bathroom, holding nothing and focused on
    nothing. There's no sign of mercury here\,\ldots''
  \item[$\vdots$]
  \item[22--26.] ``Currently, I am in the foundry, holding nothing and focused on
    the mercury\,\ldots The blast furnace is activated but still closed.''
    \hfill \emph{(identical reflection repeated for the last five turns)}
\end{enumerate}
The agent never names the active sub-goal. From turn 22 it freezes on one
reflection and emits several actions in a single turn, and the episode ends
without heating the mercury.
\end{promptbox}

\begin{promptbox}{\method{} rollout, succeeds (score 1.00)}
\small
\textit{Plan given:} 1.\ Find and obtain mercury; 2.\ Focus on the mercury;
3.\ Cause the mercury to boil or combust; 4.\ Focus on the resulting vapor or ash.
\begin{enumerate}
  \item ``\textbf{My current sub-goal is (1) Find and obtain mercury.} Currently,
    I am in the art studio\,\ldots''
  \item[$\vdots$]
  \item[11.] ``\textbf{My current sub-goal is (3) Cause the mercury to boil or
    combust.} \,\ldots I should use heat --- I see a blast furnace in the foundry.''
  \item[13.] ``\textbf{My current sub-goal is (3)}\,\ldots With the blast furnace
    activated, I should place the mercury inside it to heat it up.''
  \item[22.] ``\textbf{My current sub-goal is (3)}\,\ldots With the blast furnace
    open, I should place the mercury inside it.''
\end{enumerate}
The agent names the active sub-goal on every turn and advances through the plan.
It is not flawless, and turns 15--21 struggle with how to operate the furnace
switch, but because the reflection stays fixed on a named target it keeps
proposing \emph{different} actions toward that target instead of freezing, and it
recovers and completes the task.
\end{promptbox}

Across the run, \reflact{}+Plan's reflections restate the overall task in this
way, while \method{}'s name the active sub-goal.

\section{Additional Grounding and Plan-Quality Analyses}
\label{app:extra_analyses}

\paragraph{Invalid-Action Metric as a Grounding Proxy.}
A rejected action might be a formatting slip rather than a genuine grounding
error, so we verify four things. First, the pattern holds under
\citet{reflact2025}'s broader hallucinated-action definition, which counts
\emph{all} erroneous observations, covering ambiguous, malformed, inapplicable,
and precondition-violating actions: pooled across the three models, the rate
still orders \react{} (39.0\%) $\approx$ \reflact{} (38.8\%) above \method{}
(32.5\%) and \beammethod{} (32.3\%), with parse failures the dominant
contributor. Second, it varies sharply across backbones on the \emph{same} base
model (42.9\% for \reflact{} versus 30.3\% for \method{} on
Llama-3.1-8B-Instruct) and rises over the course of an episode---neither
pattern is explained by formatting competence, which is fixed by the base model
and does not degrade mid-episode. Third, 22--32\% of invalid actions on GPT-4o
and Llama-3.1-8B-Instruct repeat an action the environment has \emph{already}
rejected, which shows that the agent treats an infeasible action as valid rather
than mistyping a one-off command. Finally, the rise over an episode is not an
artifact of which episodes succeed: within \reflact{}'s successful episodes the
rate goes from $2.6$\% to $37.0$\%, and within its failed ones from $16.3$\% to
$76.8$\%. We nonetheless read the metric as a proxy, and leave fine-grained
annotation of syntax-versus-grounding failures to future work. Formally, for a
set $\mathcal{D}$ of executed-action observations, the rate is
$\mathrm{IA}(\mathcal{D}) = \frac{1}{|\mathcal{D}|}
 \sum_{o \in \mathcal{D}} \mathds{1}\!\left[\,o\text{ is a refusal}\,\right]$.

\paragraph{The Ordering Rule's Effect.}
We examine whether the ordering rule affects only the 36 targeted state-change
episodes or spills over into the remaining 175 (Tables~\ref{tab:procedure}
and~\ref{tab:procedure_full}). In three of the six model/planner pairs, changes
occur primarily within the targeted episodes. GPT-4o's single-path planner
gains six target episodes and loses one non-target episode, and its Beam
Search Planner gains three target and one non-target episode.
Llama-3.1-8B-Instruct's single-path planner loses two target and one
non-target episode. In the other three pairs, most of the change falls
outside the targeted episodes. GPT-4o-mini's single-path planner is the
sharpest case, gaining one target episode against nine non-target episodes
(ten additional successes overall). Llama-3.1-8B-Instruct's Beam Search
Planner gains four target and six non-target episodes, and GPT-4o-mini's
Beam Search Planner gains eleven target episodes while giving back three
non-target episodes.

\paragraph{Category Volatility and Cancellation.}
Even when the net change outside state-change tasks is near zero, sizeable
internal shifts occur. GPT-4o's single-path planner loses four \emph{Find}
episodes while gaining three \emph{Lifespan} and two \emph{Life Stages}
episodes. Llama-3.1-8B-Instruct's single-path planner loses six \emph{Find}
episodes while gaining four \emph{Measurement} and three
\emph{chemistry-mixing} episodes. Which categories move is inconsistent across
models: GPT-4o's single-path planner picks up \emph{Life Stages} (0.0 to
22.2), \emph{Lifespan} (76.7 to 86.7), and \emph{Electricity} (0.0 to 10.0),
while its Beam Search Planner shows minimal non-target movement. The sharpest
divergence is on \emph{Measurement} for GPT-4o-mini, where the single-path
planner rises from 10.0\% to 50.0\% while the Beam Search Planner falls from
45.0\% to 15.0\%; we attribute this reversal to the category's small sample
size (20 episodes), where a handful of flipped episodes moves the rate by ten
points or more at a time.

\paragraph{Interpretation and Limitations.}
These results suggest that the ordering rule acts as a general prompt for
sub-goal specificity and structure, rather than only teaching one specific
procedure. Our current ablation cannot, however, decouple whether the rule
teaches the model a specific state-change fact or improves its general
planning competence, and we leave this distinction as a question for future
work.

\begin{table}[h]
\centering
\caption{
  We report the effect of the state-change ordering rule
  (Table~\ref{tab:procedure}) on every ScienceWorld category.
  The columns are the same nine categories as Table~\ref{tab:scicat}:
  \emph{St-Chg} pools the four Boil/Freeze/Melt/Change-State task types (36
  episodes), matching Table~\ref{tab:procedure}'s State-change column, and
  \emph{Chem} pools all 26 chemistry-mixing episodes, paint and non-paint
  together.
  \emph{base} and \emph{+proc} repeat Table~\ref{tab:procedure}'s columns, and
  we bold the higher of each base/+proc pair, with ties bolded both.
}
\label{tab:procedure_full}
\scriptsize
\setlength{\tabcolsep}{3pt}
\begin{tabular}{llcccccccccc}
\toprule
Model & Planner & St-Chg & Elec & Cond & Find & Grow & Chem & Lifesp & LifeSt & Meas & All \\
\midrule
\multirow{4}{*}{GPT-4o}
& SGG base   & 55.6 & 0.0 & \textbf{50.0} & \textbf{100.0} & \textbf{10.0} & \textbf{84.6} & 76.7 & 0.0 & \textbf{90.0} & 64.0 \\
& SGG +proc  & \textbf{72.2} & \textbf{10.0} & 40.0 & 90.0 & 5.0 & \textbf{84.6} & \textbf{86.7} & \textbf{22.2} & \textbf{90.0} & \textbf{66.4} \\
& Beam base  & 66.7 & \textbf{10.0} & \textbf{50.0} & \textbf{87.5} & 0.0 & \textbf{73.1} & 80.0 & \textbf{33.3} & \textbf{90.0} & 63.5 \\
& Beam +proc & \textbf{75.0} & \textbf{10.0} & 45.0 & \textbf{87.5} & \textbf{10.0} & \textbf{73.1} & \textbf{83.3} & \textbf{33.3} & 85.0 & \textbf{65.4} \\
\midrule
\multirow{4}{*}{GPT-4o-mini}
& SGG base   & 52.8 & \textbf{0.0} & 0.0 & \textbf{95.0} & \textbf{0.0} & \textbf{65.4} & 60.0 & \textbf{11.1} & 10.0 & 45.0 \\
& SGG +proc  & \textbf{55.6} & \textbf{0.0} & \textbf{10.0} & 90.0 & \textbf{0.0} & \textbf{65.4} & \textbf{66.7} & 0.0 & \textbf{50.0} & \textbf{49.8} \\
& Beam base  & 25.0 & \textbf{0.0} & \textbf{10.0} & 85.0 & \textbf{0.0} & 61.5 & \textbf{73.3} & \textbf{0.0} & \textbf{45.0} & 43.6 \\
& Beam +proc & \textbf{55.6} & \textbf{0.0} & \textbf{10.0} & \textbf{87.5} & \textbf{0.0} & \textbf{69.2} & \textbf{73.3} & \textbf{0.0} & 15.0 & \textbf{47.4} \\
\midrule
\multirow{4}{*}{\shortstack[l]{Llama-3.1\\-8B}}
& SGG base   & \textbf{22.2} & \textbf{0.0} & 0.0 & \textbf{77.5} & \textbf{5.0} & 42.3 & \textbf{53.3} & \textbf{0.0} & 20.0 & \textbf{33.6} \\
& SGG +proc  & 16.7 & \textbf{0.0} & \textbf{10.0} & 62.5 & 0.0 & \textbf{53.8} & 43.3 & \textbf{0.0} & \textbf{40.0} & 32.2 \\
& Beam base  & 16.7 & \textbf{0.0} & \textbf{15.0} & 70.0 & 0.0 & 3.8 & \textbf{53.3} & \textbf{0.0} & 10.0 & 26.5 \\
& Beam +proc & \textbf{27.8} & \textbf{0.0} & 5.0 & \textbf{75.0} & \textbf{5.0} & \textbf{19.2} & \textbf{53.3} & \textbf{0.0} & \textbf{15.0} & \textbf{31.3} \\
\bottomrule
\end{tabular}
\end{table}

\paragraph{Search Configurations at a Comparable Budget.}
Comparison of the greedy and Beam Search Planners at a comparable budget
(Table~\ref{tab:greedy}) shows near-identical performance across most
categories: they agree exactly on six of the nine and differ by at most four
points on the rest. The main exception is chemistry mixing, where the
single-path planner (42.3\%) substantially outperforms the Beam Search
Planner (3.8\%) and its greedy variant (7.7\%). Invalid-action rates follow
the same ranking as success rates (Beam: 34.2\%, Greedy: 32.4\%,
single-path: 29.9\%). Given how little the greedy and beam variants differ,
we read them as one observation rather than two, since a single seed
separates their success rates of 27.0\% and 26.5\%.

\begin{table}[h]
\centering
\caption{
  We compare greedy versus beam search at a comparable planner budget, on
  ScienceWorld with Llama-3.1-8B-Instruct.
  The Beam Search Planner ($K=2$, $S=3$) spends at most 42 planner calls per
  episode, and its greedy variant ($K=1$, $S=6$) at most 48
  (Section~\ref{sec:method:beam}).
  All three rows use the base planner prompt, so the two rows that also
  appear in Table~\ref{tab:main} match it, and column names follow
  Table~\ref{tab:procedure_full}.
  \emph{IA} is the invalid-action rate, where lower is better; we bold the
  best result per column, with ties bolded both, and leave columns on which
  all three planners tie unmarked.
}
\label{tab:greedy}
\scriptsize
\setlength{\tabcolsep}{3pt}
\begin{tabular}{lcccccccccrc}
\toprule
Planner & St-Chg & Elec & Cond & Find & Grow & Chem & Lifesp & LifeSt & Meas & SR\,/\,AR & IA \\
\midrule
Beam ($K{=}2$, $S{=}3$)   & 16.7 & 0.0 & \textbf{15.0} & 70.0 & 0.0 & 3.8 & 53.3 & 0.0 & 10.0 & 26.5\,/\,41.9 & 34.2 \\
Greedy ($K{=}1$, $S{=}6$) & 13.9 & 0.0 & \textbf{15.0} & 72.5 & 0.0 & 7.7 & 53.3 & 0.0 & 10.0 & 27.0\,/\,41.3 & 32.4 \\
Single pass               & \textbf{22.2} & 0.0 & 0.0 & \textbf{77.5} & \textbf{5.0} & \textbf{42.3} & 53.3 & 0.0 & \textbf{20.0} & \textbf{33.6\,/\,48.7} & \textbf{29.9} \\
\bottomrule
\end{tabular}
\end{table}

\paragraph{Trajectory Length.}
To ensure \method{}'s lower invalid-action rate is not simply due to shorter
trajectories, we compare mean steps to success on shared episodes
(Table~\ref{tab:steps}), reported across all models and benchmarks on episodes
that \emph{both} \reflact{} and \method{} solve. Step counts remain comparable,
within about one step, across models, though \method{} occasionally requires
fewer (e.g., ScienceWorld on GPT-4o-mini and ALFWorld on Llama-3.1-8B-Instruct).
This indicates that \method{} succeeds by completing \emph{more} tasks rather
than merely solving them faster; results for Jericho are inconclusive given its
limited episode overlap.

\begin{table}[h]
\centering
\caption{
  We report steps to success, the mean number of environment steps, on
  episodes solved by \emph{both} \reflact{} and \method{}. $n$ is the number
  of such commonly-solved episodes.
}
\label{tab:steps}
\small
\setlength{\tabcolsep}{4pt}
\begin{tabular}{llrcc}
\toprule
Model & Benchmark & $n$ & \reflact{} & \method{} \\
\midrule
\multirow{3}{*}{GPT-4o}
& ALFWorld     & 115 & 11.2 & 11.5 \\
& ScienceWorld & 105 & 13.6 & 14.0 \\
& Jericho      &   7 & 17.0 & 19.3 \\
\midrule
\multirow{3}{*}{GPT-4o-mini}
& ALFWorld     &  78 & 10.6 & 10.7 \\
& ScienceWorld &  69 & 17.1 & 14.4 \\
& Jericho      &   2 & 13.5 & 12.5 \\
\midrule
\multirow{3}{*}{\shortstack[l]{Llama-3.1\\-8B-Instruct}}
& ALFWorld     &  35 & 10.1 &  9.0 \\
& ScienceWorld &  33 &  9.1 &  9.7 \\
& Jericho      &   1 & 23.0 &  8.0 \\
\bottomrule
\end{tabular}
\end{table}

\section{Grounding: a Hallucinated-Action Cascade (Example)}
\label{app:grounding_example}

This ScienceWorld episode (Llama-3.1-8B-Instruct, \texttt{make violet paint})
traces step by step the compounding of hallucinated actions quantified in
Section~\ref{sec:task_analysis}.
\reflact{} and \method{} face the identical task and initial state; only the
reasoning backbone differs.
The task is to use chemistry to create violet paint and then focus on it.

\begin{promptbox}{\reflact{} rollout, fails with 9 of 15 actions rejected}
\small
\begin{enumerate}
  \item \texttt{teleport to art studio} $\rightarrow$ ok
  \item \texttt{look around} $\rightarrow$ ok (paints are in cups)
  \item \texttt{open large cupboard} $\rightarrow$ ok
  \item \texttt{look at drawer} $\rightarrow$ ok
  \item \texttt{pick up cup containing yellow paint} $\rightarrow$
    \textbf{Ambiguous request}; the agent disambiguates on the next turn
  \item \texttt{1} $\rightarrow$ the wood cup moves to the inventory
  \item[7--15.] nine consecutive \texttt{pour}/\texttt{use} commands, each
    malformed and each rejected with \textbf{No known action matches that
    input}; no paint is ever mixed.
\end{enumerate}
\end{promptbox}

\reflact{} fixates on a phrasing the environment cannot parse and repeats it for
the rest of the episode, so its per-turn reflection never recovers the agent to
a valid action.

\begin{promptbox}{\method{} rollout, succeeds with 0 of 8 actions rejected}
\small
Plan: \textit{(1) teleport to the room with the materials; (2) gather the
materials; (3) mix them into violet paint; (4) focus on the result.}
\begin{enumerate}
  \item \texttt{teleport to art studio} $\rightarrow$ ok
  \item \texttt{look around} $\rightarrow$ ok
  \item \texttt{pick up wood cup containing red paint} $\rightarrow$ moved to inventory
  \item \texttt{pick up wood cup containing blue paint} $\rightarrow$ moved to inventory
  \item \texttt{pour wood cup containing blue paint into bowl} $\rightarrow$ ok
  \item \texttt{pour wood cup containing red paint into bowl} $\rightarrow$ ok
  \item \texttt{mix bowl} $\rightarrow$ ``Red and blue paint mix to produce violet paint.''
  \item \texttt{focus on violet paint} $\rightarrow$ \textbf{success}
\end{enumerate}
\end{promptbox}

The active sub-goal anchors the agent, first to \textit{gather the materials} and
then to \textit{mix them}.
\method{} therefore issues only well-formed, grounded actions and finishes in
eight steps.
Nine rejected actions versus none, on the same task, is the episode-level
signature of the aggregate invalid-action gap in Table~\ref{tab:grounding}.

\section{Actions per Episode}
\label{app:actioncount}

Table~\ref{tab:actioncount} gives the mean number of actions each agent executes
per ScienceWorld episode, over the same 211 episodes as Table~\ref{tab:main}.
It supports the argument in Section~\ref{sec:task_analysis} that the agents with
the lowest invalid-action rates are not the ones that interact least: on every
model, NoThinking and Plan-and-Act take more actions than \method{}, by $3.7$ to
$7.5$. \beammethod{} takes as few actions as \method{} or fewer, yet solves less,
which is why a low count is not on its own a sign of an efficient agent.

\begin{table}[h]
\centering
\caption{
  Mean actions executed per episode on ScienceWorld (211 episodes). Lower is not
  better on its own; read alongside the success rates in Table~\ref{tab:main} and
  the invalid-action rates in Table~\ref{tab:grounding}.
}
\label{tab:actioncount}
\small
\setlength{\tabcolsep}{6pt}
\begin{tabular}{llc}
\toprule
Model & Agent & Actions / episode \\
\midrule
\multirow{6}{*}{GPT-4o}
& NoThinking       & 24.7 \\
& \react{}         & 22.6 \\
& Plan-and-Act     & 28.3 \\
& \reflact{}       & 20.2 \\
\ourrow & \beammethod{}    & 21.6 \\
\ourrow & \method{}        & 20.8 \\
\midrule
\multirow{6}{*}{GPT-4o-mini}
& NoThinking       & 25.6 \\
& \react{}         & 23.2 \\
& Plan-and-Act     & 25.3 \\
& \reflact{}       & 23.1 \\
\ourrow & \beammethod{}    & 20.4 \\
\ourrow & \method{}        & 20.3 \\
\midrule
\multirow{6}{*}{\shortstack[l]{Llama-3.1\\-8B-Instruct}}
& NoThinking       & 22.8 \\
& \react{}         & 21.5 \\
& Plan-and-Act     & 23.4 \\
& \reflact{}       & 19.9 \\
\ourrow & \beammethod{}    & 17.5 \\
\ourrow & \method{}        & 19.1 \\
\bottomrule
\end{tabular}
\end{table}

\section{Invalid-Action Rate for All Agents}
\label{app:grounding_table}

Table~\ref{tab:grounding} gives the full invalid-action rates behind
Figure~\ref{fig:compounding}, for all six agents on all three models, overall and
by episode position.

\begin{table}[h]
\centering
\caption{
  We report the invalid-action rate (\%) on ScienceWorld for all six agents:
  the fraction of executed actions the environment refuses to interpret or
  apply, where lower is better, overall and by early/mid/late episode thirds
  (equal thirds of the actions each episode executes, pooled across episodes).
  \textbf{Bold} marks the lowest rate per model group per column; a low rate
  marks the least-rejected agent, not the best one
  (Section~\ref{sec:task_analysis}).
}
\label{tab:grounding}
\small
\begin{tabular}{llcccc}
\toprule
Model & Agent & Overall & Early & Mid & Late \\
\midrule
\multirow{6}{*}{GPT-4o}
& NoThinking     & \textbf{20.9} & 15.1 & \textbf{22.5} & 25.0 \\
& \react{}       & 35.3 & 23.1 & 41.1 & 41.7 \\
& Plan-and-Act   & 23.3 & 17.8 & 27.1 & \textbf{24.8} \\
& \reflact{}     & 34.2 & 19.0 & 40.7 & 42.9 \\
& \beammethod{}  & 24.9 & 14.6 & 30.8 & 29.3 \\
& \method{}      & 27.7 & \textbf{11.2} & 35.6 & 36.3 \\
\midrule
\multirow{6}{*}{GPT-4o-mini}
& NoThinking     & 26.6 & 21.8 & 28.5 & 29.5 \\
& \react{}       & 31.9 & 22.9 & 37.8 & 35.0 \\
& Plan-and-Act   & \textbf{24.4} & 19.7 & \textbf{28.2} & \textbf{25.4} \\
& \reflact{}     & 32.7 & 23.3 & 35.7 & 39.0 \\
& \beammethod{}  & 30.0 & 19.0 & 31.9 & 39.0 \\
& \method{}      & 32.6 & \textbf{17.7} & 34.4 & 45.5 \\
\midrule
\multirow{6}{*}{\shortstack[l]{Llama-3.1\\-8B-Instruct}}
& NoThinking     & 45.5 & 20.4 & 43.8 & 72.2 \\
& \react{}       & 44.9 & 18.6 & 47.4 & 68.7 \\
& Plan-and-Act   & 39.5 & 17.2 & 39.0 & 62.4 \\
& \reflact{}     & 43.3 & 14.5 & 44.1 & 71.4 \\
& \beammethod{}  & 34.2 & 9.3 & 36.6 & 56.9 \\
& \method{}      & \textbf{29.9} & \textbf{8.7} & \textbf{31.2} & \textbf{49.7} \\
\bottomrule
\end{tabular}
\end{table}

\section{Sub-Goal Plans: \method{} vs.\ \beammethod{} (Examples)}
\label{app:plan_examples}

The following are real plans (Llama-3.1-8B-Instruct, ScienceWorld) that
illustrate why the two planners induce different success distributions
(Section~\ref{sec:plan_quality}). Consider \texttt{melt caesium}, a task type on
which \method{} scores $0\%$ while \beammethod{} succeeds; the two plans differ
only in how the transformation is expressed.

\begin{promptbox}{\texttt{melt caesium}: \method{} (fails) vs.\ \beammethod{} (succeeds)}
\small
\textbf{\method{}} (single-path planner), \emph{fails}:
\begin{enumerate}
  \item Find and obtain caesium
  \item Focus on the caesium
  \item \textit{Take actions to increase the temperature of the caesium}
  \item \textit{Focus on the melted caesium}
\end{enumerate}
\vspace{3pt}
\textbf{\beammethod{}} (Beam Search Planner), \emph{succeeds}:
\begin{enumerate}
  \item Find and obtain the caesium
  \item Focus on the caesium
  \item \textbf{Find the heat source to melt the caesium}
  \item \textbf{Apply heat to cause the state change of the caesium}
\end{enumerate}
\end{promptbox}

Here \method{}'s plan restates the goal (``increase the temperature'') and then
passively focuses on the result, without ever naming an executable
transformation. In the corresponding rollout, the agent locates the blast
furnace but, lacking a stated ``apply heat'' step, wanders (teleport / look /
examine) and exhausts its step budget without melting the substance. \beammethod{}
instead decomposes the transformation into ``find a heat source'' and ``apply
heat,'' which the agent can directly execute.

The reverse holds on task types where \method{} already excels, which exposes the
\emph{cost} of the search. On \texttt{boil mercury}, for instance, \method{}'s
concise plan succeeds while \beammethod{}'s more elaborate one fails.

\begin{promptbox}{\texttt{boil mercury}: \method{} (succeeds) vs.\ \beammethod{} (fails)}
\small
\textbf{\method{}}, \emph{succeeds}:
\begin{enumerate}
  \item Find and obtain mercury
  \item Focus on the mercury
  \item \textbf{Cause the mercury to boil or combust}
  \item Focus on the resulting vapor or ash
\end{enumerate}
\vspace{3pt}
\textbf{\beammethod{}}, \emph{fails}:
\begin{enumerate}
  \item Find and obtain mercury
  \item \textit{Find the container holding mercury}
  \item Focus on the mercury
  \item \textit{Reach or obtain the heat source to apply to the mercury}
\end{enumerate}
\end{promptbox}

Here the search \emph{over}-decomposes: extra ``find the container'' and ``find a
heat source'' steps mislead the agent and give up the outcome that \method{}'s
direct plan attains. The two cases point the same way. What the plan says, not
whether it was found by search, decides the outcome; naming the missing operation
helps (Melt), while adding needless steps hurts (Boil).

\method{} does not always restate the goal. On \texttt{make violet paint}, a task
type where it scores highly, its plan names the operation (``\textit{Mix the
materials to create violet paint}'') and it succeeds, so what its plans supply
varies by task type. The search in \beammethod{} is itself imperfect as well: on
\texttt{freeze tin} it retrieves ``\textit{find\,\ldots\,a cold source}'' but
then instructs ``\textit{Apply heat to cause the state change},'' and this
wrong-direction step is one reason \beammethod{} still fails a majority of Freeze
episodes.


\section{Agent Prompts (ALFWorld)}
\label{app:prompts}

All prompts are reproduced verbatim from our source code, preserving all
original typographical errors. For ALFWorld, agents use the template below
with a \texttt{\{agent\_instruction\}} slot; \method{} and \beammethod{} share
the same agent instruction, differing only in how the sub-goal plan is
generated (Appendices~\ref{app:planner_prompts},~\ref{app:sci_planner},
and~\ref{app:jericho_planner}). ScienceWorld and Jericho follow analogous
templates, their action lists omitted here for space. We also reproduce
\citet{reflact2025}'s instructions as released, noting that our ScienceWorld
instruction reads ``for each of your turns \ldots from the previous turn''
against their ``for each of your turn \ldots of the last turn''
(Appendix~\ref{app:ablation_example}). Our instructions remain consistent
across ALFWorld and ScienceWorld, adding only a one-line role prefix for
Jericho to ensure the backbone is not tuned per environment. That Jericho
variant is the only instruction of ours not shown below, and it is given at
the end of this appendix.

\begin{promptbox}{Shared household prompt template}
\small
\begin{lstlisting}
Interact with a household to solve a task. Imagine you are an intelligent agent in a household environment and your target is to perform actions to complete the task goal. At the beginning of your interactions, you will be given the detailed description of the current environment and your goal to accomplish.
{agent_instruction}
The available actions are:
1. go to {recep}
2. take {obj} from {recep}
3. put {obj} in/on {recep}
4. open {recep}
5. close {recep}
6. use {obj}
7. clean {obj} with {recep}
8. heat {obj} with {recep}
9. cool {obj} with {recep}
where {obj} and {recep} correspond to objects and receptacles.
After your each turn, the environment will give you immediate feedback based on which you plan your next few steps. if the envrionment output "Nothing happened", that means the previous action is invalid and you should try more options.
Reminder:
The action must be chosen from the given available actions. Any actions except provided available actions will be regarded as illegal.
\end{lstlisting}
\end{promptbox}

Plan-and-Act uses a step-0 variant of this template that omits the ``After your each turn\ldots'' feedback sentence.

\begin{promptbox}{NoThinking instruction}
\small
\begin{lstlisting}
For each of your turn, you will be given the observation of the last turn.
You should directly output the action in this turn. Your output must strictly follow this format:"Action: your next action".
\end{lstlisting}
\end{promptbox}

\begin{promptbox}{\react{} instruction}
\small
\begin{lstlisting}
For each of your turn, you will be given the observation of the last turn. You should first think about the current condition and plan for your future actions, and then output your action in this turn. Your output must strictly follow this format:"Thought: your thoughts.
Action: your next action". Remember that you can only output one "Action:" in per response.
\end{lstlisting}
\end{promptbox}

\begin{promptbox}{Plan-and-Act instruction (step $\geq 1$)}
\small
\begin{lstlisting}
For each of your turn, you will be given the observation of the last turn.
You should directly output the action in this turn. Your output must strictly follow this format:"Action: your next action".
\end{lstlisting}
\end{promptbox}

\begin{promptbox}{Plan-and-Act planning instruction (step $0$)}
\small
\begin{lstlisting}
You should first think about the given task and plan your approach to the task, and then output the action for this turn.
\end{lstlisting}
\end{promptbox}

\begin{promptbox}{\reflact{} instruction}
\small
\begin{lstlisting}
For each of your turn, you will be given the observation of the last turn. You should first reflect in one sentence on the agent's state in relation to the task goal, and then output the action for this turn. Your output must strictly follow this format:"Reflection: your reflection.
Action: your next action".
\end{lstlisting}
\end{promptbox}

\begin{promptbox}{\method{} / \beammethod{} instruction}
\small
\begin{lstlisting}
For each of your turn, you will be given the observation of the last turn. Referring to the provided sub-goals, determine which sub-goal you are currently working on based on your progress so far. Then reflect in one sentence on your current state and progress toward the current sub-goal, and then output the action for this turn. Your output must strictly follow this format:"Reflection: your reflection.
Action: your next action".
\end{lstlisting}
\end{promptbox}

\paragraph{Jericho instruction.}\mbox{}

Our agents' instruction is unchanged from the ALFWorld one above on
ScienceWorld; on Jericho it gains the role prefix that benchmark's prompts
carry.

\begin{promptbox}{\method{} / \beammethod{} --- Jericho}
\small
\begin{lstlisting}
You are a game master in fictional text games. For each of your turn, you will be given the observation of the last turn. Referring to the provided sub-goals, determine which sub-goal you are currently working on based on your progress so far. Then reflect in one sentence on your current state and progress toward the current sub-goal, and then output the action for this turn. Your output must strictly follow this format:"Reflection: your reflection.
Action: your next action".
\end{lstlisting}
\end{promptbox}

\section{\method{} In-Context Examples}
\label{app:icl}

The single-shot in-context example given to the \method{} agent (shared with \beammethod{}), one per environment; ALFWorld provides one example per task type.

\begin{promptbox}{ALFWorld --- \texttt{pick\_and\_place}}
\small
\begin{lstlisting}
USER: You are in the middle of a room. Looking quickly around you, you see a cabinet 4, a cabinet 3, a cabinet 2, a cabinet 1, a countertop 1, a garbagecan 1, a handtowelholder 2, a handtowelholder 1, a sinkbasin 2, a sinkbasin 1, a toilet 1, a toiletpaperhanger 1, and a towelholder 1.
Your task is to: put some spraybottle on toilet.

USER: Here is your sub goals to complete the task:
1. Locate the required object
2. Obtain the required object
3. Bring the object to the target receptacle
4. Place the object to complete the task

ASSISTANT: Reflection: My current sub-goal is Locate the required object. Currently, I am in the middle of the room, not holding anything. A spraybottle is likely in the cabinets -- I should check cabinet 1 first.
Action: go to cabinet 1

USER: Observation: On the cabinet 1, you see a cloth 1, a soapbar 1, a soapbottle 1.

ASSISTANT: Reflection: My current sub-goal is Locate the required object. Currently, I am at cabinet 1, not holding anything. There is no spraybottle here -- I should check the next cabinet.
Action: go to cabinet 2

USER: Observation: The cabinet 2 is closed.

ASSISTANT: Reflection: My current sub-goal is Locate the required object. Currently, I am at cabinet 2, not holding anything. The cabinet is closed -- I need to open it to look inside.
Action: open cabinet 2

USER: Observation: You open the cabinet 2. The cabinet 2 is open. In it, you see a candle 1, and a spraybottle 2.

ASSISTANT: Reflection: The sub-goal is complete -- I found a spraybottle 2 in cabinet 2. My current sub-goal is Obtain the required object. Currently, I am at cabinet 2, not holding anything -- I should take the spraybottle now.
Action: take spraybottle 2 from cabinet 2

USER: Observation: You pick up the spraybottle 2 from the cabinet 2.

ASSISTANT: Reflection: The sub-goal is complete -- I now hold the spraybottle 2. My current sub-goal is Bring the object to the target receptacle. Currently, I am at cabinet 2, holding the spraybottle -- I should go to the toilet, the target receptacle.
Action: go to toilet 1

USER: Observation: On the toilet 1, you see a soapbottle 2.

ASSISTANT: Reflection: The sub-goal is complete -- I have arrived at toilet 1 with the spraybottle. My current sub-goal is Place the object to complete the task. Currently, I am at toilet 1, holding the spraybottle -- I should place it now.
Action: put spraybottle 2 in/on toilet 1

USER: Observation: You put the spraybottle 2 in/on the toilet 1.
\end{lstlisting}
\end{promptbox}

\begin{promptbox}{ALFWorld --- \texttt{pick\_clean\_then\_place}}
\small
\begin{lstlisting}
USER: You are in the middle of a room. Looking quickly around you, you see a cabinet 13, a cabinet 12, a cabinet 11, a cabinet 10, a cabinet 9, a cabinet 8, a cabinet 7, a cabinet 6, a cabinet 5, a cabinet 4, a cabinet 3, a cabinet 2, a cabinet 1, a coffeemachine 1, a countertop 1, a diningtable 1, a drawer 1, a fridge 1, a garbagecan 1, a microwave 1, a shelf 3, a shelf 2, a shelf 1, a sinkbasin 1, a stoveburner 4, a stoveburner 3, a stoveburner 2, a stoveburner 1, and a toaster 1.
Your task is to: put a clean lettuce in diningtable.

USER: Here is your sub goals to complete the task:
1. Find and obtain the required object
2. Clean the object at the cleaning station
3. Bring the cleaned object to the target receptacle
4. Place the object to complete the task

ASSISTANT: Reflection: My current sub-goal is Find and obtain the required object. Currently, I am in the middle of the room, not holding anything. Lettuce is often kept cold, so I should check the fridge first.
Action: go to fridge 1

USER: Observation: The fridge 1 is closed.

ASSISTANT: Reflection: My current sub-goal is Find and obtain the required object. Currently, I am at fridge 1, not holding anything. The fridge is closed -- I need to open it to look inside.
Action: open fridge 1

USER: Observation: You open the fridge 1. The fridge 1 is open. In it, you see a cup 3, a egg 2, a potato 3, and a potato 2.

ASSISTANT: Reflection: My current sub-goal is Find and obtain the required object. Currently, I am at fridge 1, not holding anything. There is no lettuce in the fridge -- I should check the diningtable next.
Action: go to diningtable 1

USER: Observation: On the diningtable 1, you see a apple 1, a bread 1, a butterknife 2, a cup 2, a fork 2, a knife 2, a knife 1, a ladle 1, a lettuce 1, a mug 2, a mug 1, a pan 2, a peppershaker 1, a spatula 3, a tomato 2, and a tomato 1.

ASSISTANT: Reflection: My current sub-goal is Find and obtain the required object. Currently, I am at diningtable 1, not holding anything. I see the lettuce 1 here -- I should take it.
Action: take lettuce 1 from diningtable 1

USER: Observation: You pick up the lettuce 1 from the diningtable 1.

ASSISTANT: Reflection: The sub-goal is complete -- I now hold the lettuce 1. My current sub-goal is Clean the object at the cleaning station. Currently, I am at diningtable 1, holding the lettuce -- I should go to the sinkbasin to clean it.
Action: go to sinkbasin 1

USER: Observation: On the sinkbasin 1, you see a apple 2, a ladle 2, a spoon 1, and a tomato 3.

ASSISTANT: Reflection: My current sub-goal is Clean the object at the cleaning station. Currently, I am at sinkbasin 1, holding the lettuce -- I should clean it now.
Action: clean lettuce 1 with sinkbasin 1

USER: Observation: You clean the lettuce 1 using the sinkbasin 1.

ASSISTANT: Reflection: The sub-goal is complete -- the lettuce is now clean. My current sub-goal is Bring the cleaned object to the target receptacle. Currently, I am at sinkbasin 1, holding the clean lettuce -- I should go to the diningtable.
Action: go to diningtable 1

USER: Observation: On the diningtable 1, you see a apple 1, a bread 1, a butterknife 2, a cup 2, a fork 2, a knife 2, a knife 1, a ladle 1, a mug 2, a mug 1, a pan 2, a peppershaker 1, a spatula 3, a tomato 2, and a tomato 1.

ASSISTANT: Reflection: The sub-goal is complete -- I have arrived at diningtable 1 with the clean lettuce. My current sub-goal is Place the object to complete the task. Currently, I am at diningtable 1, holding the clean lettuce -- I should place it now.
Action: put lettuce 1 in/on diningtable 1

USER: Observation: You put the lettuce 1 in/on the diningtable 1.
\end{lstlisting}
\end{promptbox}

\begin{promptbox}{ALFWorld --- \texttt{pick\_heat\_then\_place}}
\small
\begin{lstlisting}
USER: You are in the middle of a room. Looking quickly around you, you see a cabinet 10, a cabinet 9, a cabinet 8, a cabinet 7, a cabinet 6, a cabinet 5, a cabinet 4, a cabinet 3, a cabinet 2, a cabinet 1, a coffeemachine 1, a countertop 3, a countertop 2, a countertop 1, a diningtable 1, a drawer 6, a drawer 5, a drawer 4, a drawer 3, a drawer 2, a drawer 1, a fridge 1, a garbagecan 1, a microwave 1, a sinkbasin 1, a stoveburner 4, a stoveburner 3, a stoveburner 2, a stoveburner 1, and a toaster 1.
Your task is to: heat some egg and put it in diningtable.

USER: Here is your sub goals to complete the task:
1. Find and obtain the required object
2. Heat the object at the heating appliance
3. Bring the heated object to the target receptacle
4. Place the object to complete the task

ASSISTANT: Reflection: My current sub-goal is Find and obtain the required object. Currently, I am in the middle of the room, not holding anything. An egg might be in the fridge -- I should open it to check.
Action: open fridge 1

USER: Observation: You open the fridge 1. The fridge 1 is open. In it, you see a lettuce 2, a mug 2, and a potato 3.

ASSISTANT: Reflection: My current sub-goal is Find and obtain the required object. Currently, I am at fridge 1, not holding anything. There is no egg in the fridge -- I should check the countertops.
Action: go to countertop 1

USER: Observation: On the countertop 1, you see a bread 1, a fork 1, and a saltshaker 1.

ASSISTANT: Reflection: My current sub-goal is Find and obtain the required object. Currently, I am at countertop 1, not holding anything. No egg here -- I should check another countertop.
Action: go to countertop 3

USER: Observation: On the countertop 3, you see a bowl 1, a butterknife 1, a egg 2, a kettle 2, a plate 1, a sink 1, and a spatula 2.

ASSISTANT: Reflection: My current sub-goal is Find and obtain the required object. Currently, I am at countertop 3, not holding anything. I see the egg 2 here -- I should take it.
Action: take egg 2 from countertop 3

USER: Observation: You pick up the egg 2 from the countertop 3.

ASSISTANT: Reflection: The sub-goal is complete -- I now hold the egg 2. My current sub-goal is Heat the object at the heating appliance. Currently, I am at countertop 3, holding the egg -- I should go to the microwave to heat it.
Action: go to microwave 1

USER: Observation: The microwave 1 is closed.

ASSISTANT: Reflection: My current sub-goal is Heat the object at the heating appliance. Currently, I am at microwave 1, holding the egg -- I should heat the egg with the microwave now.
Action: heat egg 2 with microwave 1

USER: Observation: You heat the egg 2 using the microwave 1.

ASSISTANT: Reflection: The sub-goal is complete -- the egg is now heated. My current sub-goal is Bring the heated object to the target receptacle. Currently, I am at microwave 1, holding the heated egg -- I should go to the diningtable.
Action: go to diningtable 1

USER: Observation: On the diningtable 1, you see a apple 2, a bread 3, a egg 1, a kettle 1, a knife 1, a mug 1, a papertowelroll 1, a peppershaker 2, a potato 1, a soapbottle 1, and a spatula 1.

ASSISTANT: Reflection: The sub-goal is complete -- I have arrived at diningtable 1 with the heated egg. My current sub-goal is Place the object to complete the task. Currently, I am at diningtable 1, holding the heated egg -- I should place it now.
Action: put egg 2 in/on diningtable 1

USER: Observation: You put the egg 2 in/on the diningtable 1.
\end{lstlisting}
\end{promptbox}

\begin{promptbox}{ALFWorld --- \texttt{pick\_cool\_then\_place}}
\small
\begin{lstlisting}
USER: You are in the middle of a room. Looking quickly around you, you see a cabinet 16, a cabinet 15, a cabinet 14, a cabinet 13, a cabinet 12, a cabinet 11, a cabinet 10, a cabinet 9, a cabinet 8, a cabinet 7, a cabinet 6, a cabinet 5, a cabinet 4, a cabinet 3, a cabinet 2, a cabinet 1, a coffeemachine 1, a countertop 2, a countertop 1, a diningtable 1, a drawer 5, a drawer 4, a drawer 3, a drawer 2, a drawer 1, a fridge 1, a garbagecan 1, a microwave 1, a safe 1, a sinkbasin 1, a stoveburner 4, a stoveburner 3, a stoveburner 2, a stoveburner 1, and a toaster 1.
Your task is to: cool some pan and put it in stoveburner.

USER: Here is your sub goals to complete the task:
1. Find and obtain the required object
2. Cool the object at the cooling appliance
3. Bring the cooled object to the target receptacle
4. Place the object to complete the task

ASSISTANT: Reflection: My current sub-goal is Find and obtain the required object. Currently, I am in the middle of the room, not holding anything. A pan is often on a stoveburner -- I should check stoveburner 1.
Action: go to stoveburner 1

USER: Observation: On the stoveburner 1, you see nothing.

ASSISTANT: Reflection: My current sub-goal is Find and obtain the required object. Currently, I am at stoveburner 1, not holding anything. Nothing here -- I should check another stoveburner.
Action: go to stoveburner 3

USER: Observation: On the stoveburner 3, you see a pan 1.

ASSISTANT: Reflection: My current sub-goal is Find and obtain the required object. Currently, I am at stoveburner 3, not holding anything. I see the pan 1 here -- I should take it.
Action: take pan 1 from stoveburner 3

USER: Observation: You pick up the pan 1 from the stoveburner 3.

ASSISTANT: Reflection: The sub-goal is complete -- I now hold the pan 1. My current sub-goal is Cool the object at the cooling appliance. Currently, I am at stoveburner 3, holding the pan -- I should go to the fridge to cool it.
Action: go to fridge 1

USER: Observation: The fridge 1 is closed.

ASSISTANT: Reflection: My current sub-goal is Cool the object at the cooling appliance. Currently, I am at fridge 1, holding the pan -- I should cool the pan with the fridge now.
Action: cool pan 1 with fridge 1

USER: Observation: You cool the pan 1 using the fridge 1.

ASSISTANT: Reflection: The sub-goal is complete -- the pan is now cooled. My current sub-goal is Bring the cooled object to the target receptacle. Currently, I am at fridge 1, holding the cooled pan -- I should go to the stoveburner.
Action: go to stoveburner 1

USER: Observation: On the stoveburner 1, you see nothing.

ASSISTANT: Reflection: The sub-goal is complete -- I have arrived at stoveburner 1 with the cooled pan. My current sub-goal is Place the object to complete the task. Currently, I am at stoveburner 1, holding the cooled pan -- I should place it now.
Action: put pan 1 in/on stoveburner 1

USER: Observation: You put the pan 1 in/on the stoveburner 1.
\end{lstlisting}
\end{promptbox}

\begin{promptbox}{ALFWorld --- \texttt{pick\_two\_obj}}
\small
\begin{lstlisting}
USER: You are in the middle of a room. Looking quickly around you, you see a armchair 2, a armchair 1, a bed 1, a countertop 1, a diningtable 1, a drawer 2, a drawer 1, a dresser 1, a garbagecan 1, a laundryhamper 1, and a sidetable 1.
Your task is to: put two creditcard in dresser.

USER: Here is your sub goals to complete the task:
1. Find and obtain the first required object
2. Bring the first object to the target receptacle and place it
3. Find and obtain the second required object
4. Bring the second object to the target receptacle and place it

ASSISTANT: Reflection: My current sub-goal is Find and obtain the first required object. Currently, I am in the middle of the room, not holding anything. A creditcard might be in a drawer -- I should check drawer 1.
Action: go to drawer 1

USER: Observation: The drawer 1 is closed.

ASSISTANT: Reflection: My current sub-goal is Find and obtain the first required object. Currently, I am at drawer 1, not holding anything. The drawer is closed -- I need to open it.
Action: open drawer 1

USER: Observation: You open the drawer 1. The drawer 1 is open. In it, you see a book 1, a cd 1, and a pen 1.

ASSISTANT: Reflection: My current sub-goal is Find and obtain the first required object. Currently, I am at drawer 1, not holding anything. No creditcard here -- I should check the countertop.
Action: go to countertop 1

USER: Observation: On the countertop 1, you see a cellphone 2, a creditcard 4, a creditcard 3, a creditcard 2, a mirror 1, a pencil 2, and a pencil 1.

ASSISTANT: Reflection: My current sub-goal is Find and obtain the first required object. Currently, I am at countertop 1, not holding anything. I see creditcards here -- I should take the first one.
Action: take creditcard 2 from countertop 1

USER: Observation: You pick up the creditcard 2 from the countertop 1.

ASSISTANT: Reflection: The sub-goal is complete -- I now hold the first creditcard. My current sub-goal is Bring the first object to the target receptacle and place it. Currently, I am at countertop 1, holding creditcard 2 -- I should go to the dresser.
Action: go to dresser 1

USER: Observation: On the dresser 1, you see a mug 1, and a television 1.

ASSISTANT: Reflection: My current sub-goal is Bring the first object to the target receptacle and place it. Currently, I am at dresser 1, holding creditcard 2 -- I should place it now.
Action: put creditcard 2 in/on dresser 1

USER: Observation: You put the creditcard 2 in/on the dresser 1.

ASSISTANT: Reflection: The sub-goal is complete -- the first creditcard is in the dresser. My current sub-goal is Find and obtain the second required object. Currently, I am at dresser 1, not holding anything -- I should return to the countertop for the second creditcard.
Action: go to countertop 1

USER: Observation: On the countertop 1, you see a cellphone 2, a creditcard 4, a creditcard 3, a mirror 1, a pencil 2, and a pencil 1.

ASSISTANT: Reflection: My current sub-goal is Find and obtain the second required object. Currently, I am at countertop 1, not holding anything. I see another creditcard here -- I should take it.
Action: take creditcard 3 from countertop 1

USER: Observation: You pick up the creditcard 3 from the countertop 1.

ASSISTANT: Reflection: The sub-goal is complete -- I now hold the second creditcard. My current sub-goal is Bring the second object to the target receptacle and place it. Currently, I am at countertop 1, holding creditcard 3 -- I should go to the dresser.
Action: go to dresser 1

USER: Observation: On the dresser 1, you see a creditcard 2, a mug 1, and a television 1.

ASSISTANT: Reflection: My current sub-goal is Bring the second object to the target receptacle and place it. Currently, I am at dresser 1, holding creditcard 3 -- I should place it now to complete the task.
Action: put creditcard 3 in/on dresser 1

USER: Observation: You put the creditcard 3 in/on the dresser 1.
\end{lstlisting}
\end{promptbox}

\begin{promptbox}{ALFWorld --- \texttt{look\_at\_obj}}
\small
\begin{lstlisting}
USER: You are in the middle of a room. Looking quickly around you, you see a bed 1, a desk 1, a drawer 3, a drawer 2, a drawer 1, a garbagecan 1, a safe 1, a shelf 5, a shelf 4, a shelf 3, a shelf 2, a shelf 1, a sidetable 2, and a sidetable 1.
Your task is to: look at bowl under the desklamp.

USER: Here is your sub goals to complete the task:
1. Locate the required object
2. Obtain the required object
3. Locate the light source for examination
4. Examine the object under the light source to complete the task

ASSISTANT: Reflection: My current sub-goal is Locate the required object. Currently, I am in the middle of the room, not holding anything. A bowl might be on a shelf -- I should check shelf 1.
Action: go to shelf 1

USER: Observation: On the shelf 1, you see nothing.

ASSISTANT: Reflection: My current sub-goal is Locate the required object. Currently, I am at shelf 1, not holding anything. Nothing here -- I should check shelf 2.
Action: go to shelf 2

USER: Observation: On the shelf 2, you see a bowl 1.

ASSISTANT: Reflection: The sub-goal is complete -- I found the bowl 1 on shelf 2. My current sub-goal is Obtain the required object. Currently, I am at shelf 2, not holding anything -- I should take the bowl.
Action: take bowl 1 from shelf 2

USER: Observation: You pick up the bowl 1 from the shelf 2.

ASSISTANT: Reflection: The sub-goal is complete -- I now hold the bowl 1. My current sub-goal is Locate the light source for examination. Currently, I am at shelf 2, holding the bowl -- a desklamp is often on a sidetable, so I should check sidetable 1.
Action: go to sidetable 1

USER: Observation: On the sidetable 1, you see a cd 1, a pen 1, and a pencil 1.

ASSISTANT: Reflection: My current sub-goal is Locate the light source for examination. Currently, I am at sidetable 1, holding the bowl. No desklamp here -- I should check sidetable 2.
Action: go to sidetable 2

USER: Observation: On the sidetable 2, you see a alarmclock 1, a desklamp 1, and a pen 2.

ASSISTANT: Reflection: The sub-goal is complete -- I found the desklamp 1 on sidetable 2. My current sub-goal is Examine the object under the light source to complete the task. Currently, I am at sidetable 2, holding the bowl -- I should use the desklamp to examine it.
Action: use desklamp 1

USER: Observation: You turn on the desklamp 1.
\end{lstlisting}
\end{promptbox}

\begin{promptbox}{ScienceWorld}
\small
\begin{lstlisting}
USER: Your task is to use chemistry to create green paint. When you are done, focus on the green paint.

USER: Here is your sub goals to complete the task:
1. Teleport to the room where the needed materials are found
2. Gather the materials needed for green paint
3. Mix the materials to create green paint
4. Focus on the resulting green paint

ASSISTANT: Reflection: My current sub-goal is Teleport to the room where the needed materials are found. Currently, I am in the starting room, holding nothing and focused on nothing. Paint materials are likely to be found in the art studio -- I should teleport there immediately to begin gathering.
Action: teleport to art studio

USER: Observation: You teleport to the art studio.

ASSISTANT: Reflection: The sub-goal is complete -- I have arrived at the art studio. My current sub-goal is Gather the materials needed for green paint. Currently, I am in the art studio, holding nothing and focused on nothing. I need to identify what materials are available here -- I should look around first.
Action: look around

USER: Observation: This room is called the art studio. In it, you see: 
  the agent
  a substance called air
  a large cupboard. The large cupboard door is open. In the large cupboard is: a drawer, a wood cup (containing yellow paint), a wood cup (containing blue paint), a wood cup (containing red paint).
  a table. On the table is: a glass cup (containing nothing).
  a wood cup (containing yellow paint)
  a wood cup (containing blue paint)
  a wood cup (containing red paint)
You also see:
  A door to the hallway (that is open)

ASSISTANT: Reflection: My current sub-goal is Gather the materials needed for green paint. Currently, I am in the art studio, holding nothing and focused on nothing. Blue and yellow make green, and both are available here -- I should pick up the blue paint first to start gathering the required materials.
Action: pick up wood cup containing blue paint

USER: Observation: You move the wood cup to the inventory.

ASSISTANT: Reflection: My current sub-goal is Gather the materials needed for green paint. Currently, I am in the art studio, holding the blue paint and focused on nothing. I have the blue paint -- I now need to pick up the yellow paint to complete gathering both required materials.
Action: pick up wood cup containing yellow paint

USER: Observation: You move the wood cup to the inventory.

ASSISTANT: Reflection: The sub-goal is complete -- I now hold both blue and yellow paint. My current sub-goal is Mix the materials to create green paint. Currently, I am in the art studio, holding blue and yellow paint and focused on nothing. I need to start mixing -- I should pour the blue paint into the empty glass cup first.
Action: pour wood cup containing blue paint into glass cup

USER: Observation: You pour the contents of the wood cup into the glass cup.

ASSISTANT: Reflection: My current sub-goal is Mix the materials to create green paint. Currently, I am in the art studio, holding the yellow paint and focused on nothing. The blue paint is now in the glass cup -- I need to add the yellow paint now to have both colors together for mixing.
Action: pour wood cup containing yellow paint into glass cup

USER: Observation: You pour the contents of the wood cup into the glass cup.

ASSISTANT: Reflection: My current sub-goal is Mix the materials to create green paint. Currently, I am in the art studio, holding nothing and focused on nothing. Both blue and yellow paint are now in the glass cup -- I need to mix them now to produce the green paint.
Action: mix glass cup

USER: Observation: Blue and yellow paint mix to produce green paint.

ASSISTANT: Reflection: The sub-goal is complete -- the green paint has been created. My current sub-goal is Focus on the resulting green paint. Currently, I am in the art studio, holding nothing and focused on nothing. The green paint is ready -- I need to focus on it now to complete the task.
Action: focus on green paint

USER: Observation: You focus on the green paint.
\end{lstlisting}
\end{promptbox}

\begin{promptbox}{Jericho}
\small
\begin{lstlisting}
USER: Goal: You are the warrior Link that needs to save the princess from the castle.
Observation: You are at the path leading to the castle. The castle is to your north. There is a barrel in front of you.

USER: Here is your sub goals to complete the task:
1. Find and obtain any useful item available in the starting area
2. Navigate into the castle
3. Reach the princess's location within the castle
4. Wake the princess to complete the rescue

ASSISTANT: Reflection: My current sub-goal is Find and obtain any useful item available in the starting area. Currently, I am at the path leading to the castle, holding nothing. There is a barrel in front of me -- I should examine it to find any useful item before heading into the castle.
Action: examine barrel

USER: Observation: There is a gleaming elvish sword.

ASSISTANT: Reflection: My current sub-goal is Find and obtain any useful item available in the starting area. Currently, I am at the path leading to the castle, holding nothing. I found a gleaming elvish sword in the barrel -- I need to take it now to be equipped before entering the castle.
Action: take the sword

USER: Observation: The sword is taken

ASSISTANT: Reflection: The sub-goal is complete -- I now hold the elvish sword. My current sub-goal is Navigate into the castle. Currently, I am at the path, holding the elvish sword. The castle entrance is to the north -- I should move there now.
Action: north

USER: Observation: Main hall
The castle is dimly lit, with lava flowing in many places. There is a staircase leading up to princess bedroom, a door to your north leading to the kitchen hall, and the door you just entered on your south

ASSISTANT: Reflection: The sub-goal is complete -- I have entered the castle. My current sub-goal is Reach the princess's location within the castle. Currently, I am in the castle main hall, holding the elvish sword. A staircase leads up to the princess's bedroom -- I should check the valid actions to confirm the correct path.
Action: check valid actions

USER: Observation: south, north, up, look, examine staircase

ASSISTANT: Reflection: My current sub-goal is Reach the princess's location within the castle. Currently, I am in the castle main hall, holding the elvish sword. The valid actions confirm 'up' leads to the princess's bedroom -- I should go up now to reach her.
Action: up

USER: Observation: Princess's bedroom. Princess is sleeping in bed.

ASSISTANT: Reflection: The sub-goal is complete -- I have reached the princess's bedroom. My current sub-goal is Wake the princess to complete the rescue. Currently, I am in the princess's bedroom, holding the elvish sword, and the princess is sleeping -- I should check the valid actions to find how to wake her.
Action: check valid actions

USER: Observation: wake up the princess, take out sword, down

ASSISTANT: Reflection: My current sub-goal is Wake the princess to complete the rescue. Currently, I am in the princess's bedroom, holding the elvish sword. The action "wake up the princess" is available and directly achieves this final sub-goal -- I should use it now.
Action: wake up the princess

USER: Observation: The princess wake up from the coma. Thank you my knight, she says. The task is finished.
\end{lstlisting}
\end{promptbox}

\section{ALFWorld Planner Prompts and In-Context Example}
\label{app:planner_prompts}

The \method{} planner for ALFWorld produces phase-level sub-goals. We show a representative in-context example (\texttt{pick\_and\_place}). The other five task types follow the same format.

\begin{promptbox}{One-shot planner prompt}
\small
\begin{lstlisting}
Generate exactly 4 high-level sub-goals for a task in a text-based household simulation environment (ALFWorld). Each sub-goal is an outcome or milestone -- WHAT to achieve, not HOW -- and may require multiple actions; do NOT specify exact action commands. Rules: (1) Order sub-goals by dependency: the object must be found and obtained before it is transformed (cleaned, heated, or cooled) or placed. (2) If the task requires a transformation (clean/heat/cool), the transformation sub-goal MUST come before the placement sub-goal. (3) Use generic descriptions for variable elements (e.g., 'the required object', 'the target receptacle', 'the appropriate appliance'); only name objects or locations explicitly stated in the task. (4) The final sub-goal must place or use the object to directly achieve the task goal. (5) Write each sub-goal as a short imperative phrase starting with a verb, with no trailing period. When an object is named in the task, weave it inline (e.g., 'Find the peppershaker'), not as a trailing clause. Format: 'Sub goals:\n1. ...\n2. ...\n...'
\end{lstlisting}
\end{promptbox}

\begin{promptbox}{One-shot planner in-context example}
\small
\begin{lstlisting}
USER: You are in the middle of a room. Looking quickly around you, you see a cabinet 4, a cabinet 3, a cabinet 2, a cabinet 1, a countertop 1, a garbagecan 1, a handtowelholder 2, a handtowelholder 1, a sinkbasin 2, a sinkbasin 1, a toilet 1, a toiletpaperhanger 1, and a towelholder 1.
Your task is to: put some spraybottle on toilet.

ASSISTANT: Sub goals:
1. Locate the required object
2. Obtain the required object
3. Bring the object to the target receptacle
4. Place the object to complete the task
\end{lstlisting}
\end{promptbox}

\begin{promptbox}{Beam Search Planner prompt (per-step candidate generation)}
\small
\begin{lstlisting}
Generate the next high-level sub-goal to extend a partial plan for a task in a text-based household simulation environment (ALFWorld). Each sub-goal is an outcome or milestone -- WHAT to achieve, not HOW -- and may require multiple actions; do NOT specify exact action commands. Rules: (1) Order by dependency: the object must be found and obtained before it is transformed (cleaned, heated, or cooled) or placed; do not depend on something a later sub-goal will provide. (2) If the task requires a transformation (clean/heat/cool), the transformation sub-goal MUST come before the placement sub-goal. (3) Use generic descriptions for variable elements (e.g., 'the required object', 'the target receptacle', 'the appropriate appliance'); only name objects or locations explicitly stated in the task. (4) If this is the 4th (last) sub-goal, it must place or use the object to directly achieve the task goal. (5) Write the sub-goal as a short imperative phrase starting with a verb, with no trailing period. When an object is named in the task, weave it inline (e.g., 'Find the peppershaker'), not as a trailing clause. The total plan must consist of exactly 4 sub-goals. Output ONLY the sub-goal description on a single line. No explanation, no prefix.
\end{lstlisting}
\end{promptbox}

Here \texttt{\{expand\_prompt\}} is the Beam Search Planner prompt above.

\begin{promptbox}{Self-evaluation prompt}
\small
\begin{lstlisting}
Evaluate whether a partial plan is appropriate given the following planning constraints:

{expand_prompt}

Given the Task and its Partial plan, assess how well the plan satisfies ALL of the following:
  1. Each sub-goal describes WHAT to achieve (an outcome or milestone), not HOW -- it may require multiple actions.
  2. Each sub-goal logically follows from the previous ones without skipping prerequisites.
  3. An object must be located and obtained before any sub-goal that transforms (cleans, heats, cools) or places it.
  4. Sub-goals use generic descriptions for variable elements (e.g., 'the required object', 'the target receptacle', 'the available cleaning station').
  5. If it is the 4th (last) sub-goal, completing it achieves the final goal stated in the Task.
Rate how well the Partial plan satisfies the criteria above on a continuous scale from 0.0 to 1.0 (1.0 = fully correct and well-ordered, 0.0 = clearly wrong). Use the full range: a plan that satisfies most but not all criteria should receive an intermediate value.
Answer with a single decimal number only (e.g., 0.82). No explanation.
\end{lstlisting}
\end{promptbox}

\begin{promptbox}{Beam Search Planner in-context example}
\small
\begin{lstlisting}
USER: Task: You are in the middle of a room. Looking quickly around you, you see a cabinet 4, a cabinet 3, a cabinet 2, a cabinet 1, a countertop 1, a garbagecan 1, a handtowelholder 2, a handtowelholder 1, a sinkbasin 2, a sinkbasin 1, a toilet 1, a toiletpaperhanger 1, and a towelholder 1.
Your task is to: put some spraybottle on toilet.
Current plan so far: []
Generate the next sub-goal (step 1 of 4):

ASSISTANT: Locate the required object

USER: Task: You are in the middle of a room. Looking quickly around you, you see a cabinet 4, a cabinet 3, a cabinet 2, a cabinet 1, a countertop 1, a garbagecan 1, a handtowelholder 2, a handtowelholder 1, a sinkbasin 2, a sinkbasin 1, a toilet 1, a toiletpaperhanger 1, and a towelholder 1.
Your task is to: put some spraybottle on toilet.
Current plan so far: ['Locate the required object']
Generate the next sub-goal (step 2 of 4):

ASSISTANT: Obtain the required object

USER: Task: You are in the middle of a room. Looking quickly around you, you see a cabinet 4, a cabinet 3, a cabinet 2, a cabinet 1, a countertop 1, a garbagecan 1, a handtowelholder 2, a handtowelholder 1, a sinkbasin 2, a sinkbasin 1, a toilet 1, a toiletpaperhanger 1, and a towelholder 1.
Your task is to: put some spraybottle on toilet.
Current plan so far: ['Locate the required object', 'Obtain the required object']
Generate the next sub-goal (step 3 of 4):

ASSISTANT: Bring the object to the target receptacle

USER: Task: You are in the middle of a room. Looking quickly around you, you see a cabinet 4, a cabinet 3, a cabinet 2, a cabinet 1, a countertop 1, a garbagecan 1, a handtowelholder 2, a handtowelholder 1, a sinkbasin 2, a sinkbasin 1, a toilet 1, a toiletpaperhanger 1, and a towelholder 1.
Your task is to: put some spraybottle on toilet.
Current plan so far: ['Locate the required object', 'Obtain the required object', 'Bring the object to the target receptacle']
Generate the next sub-goal (step 4 of 4):

ASSISTANT: Place the object to complete the task
\end{lstlisting}
\end{promptbox}

\section{ScienceWorld Planner Prompts and In-Context Example}
\label{app:sci_planner}

\begin{promptbox}{One-shot planner prompt}
\small
\begin{lstlisting}
Generate exactly 4 high-level sub-goals to help an agent complete a task in a text-based science simulation environment (ScienceWorld). Each sub-goal describes WHAT to achieve, not HOW -- it may require multiple actions. Rules: (1) Follow ordering instructions in the task description exactly. CRITICAL: if the task says 'First, focus on the substance' (or 'first focus on X'), the 'focus on' sub-goal MUST come immediately after finding/obtaining the substance (around sub-goal 2) and BEFORE any sub-goal that changes, heats, cools, boils, or otherwise transforms it. NEVER place such a 'focus on' sub-goal last. (2) Any 'focus on' sub-goal must appear after the sub-goal that finds or obtains the object. (3) Use generic descriptions for variable elements (e.g., 'the room where X can be found', 'the container holding X'). Only name objects or substances explicitly stated in the task. Format: 'Sub goals:\n1. ...\n2. ...\n...'
\end{lstlisting}
\end{promptbox}

\begin{promptbox}{One-shot planner in-context example}
\small
\begin{lstlisting}
USER: Your task is to use chemistry to create green paint. When you are done, focus on the green paint.

ASSISTANT: Sub goals:
1. Teleport to the room where the needed materials are found
2. Gather the materials needed for green paint
3. Mix the materials to create green paint
4. Focus on the resulting green paint
\end{lstlisting}
\end{promptbox}

\begin{promptbox}{Beam Search Planner prompt (per-step candidate generation)}
\small
\begin{lstlisting}
Generate the next candidate sub-goal to extend a partial plan for a task in a text-based science simulation environment (ScienceWorld). Each sub-goal describes WHAT to achieve, not HOW -- it may require multiple actions. Rules: (1) The sub-goal content MUST be derived from the task description, not from any example. If the task description explicitly states what to do next (e.g., 'First, focus on the substance', 'First, focus on the thing'), that action MUST become the next sub-goal -- expressed without the 'First,' prefix. (2) A 'focus on' sub-goal MUST appear after finding the object AND before moving or placing it. CRITICAL: Even if the task description says 'First, focus on X', when the current plan is empty, the first sub-goal MUST be finding or locating X -- never focusing. (3) Use generic descriptions for variable elements (e.g., 'the room where X can be found', 'the container holding X'). Only name substances or objects explicitly stated in the task. The total plan must consist of exactly 4 sub-goals. Output ONLY the sub-goal description on a single line. No explanation, no prefix.
\end{lstlisting}
\end{promptbox}

Here \texttt{\{expand\_prompt\}} is the Beam Search Planner prompt above.

\begin{promptbox}{Self-evaluation prompt}
\small
\begin{lstlisting}
Evaluate whether a partial plan is appropriate given the following planning constraints:

{expand_prompt}

Given the Task and its Partial plan, assess how well the plan satisfies ALL of the following:
  1. Each sub-goal describes WHAT to achieve (not HOW) and follows any ordering instructions in the task description.
  2. Each sub-goal logically follows from the previous ones without skipping prerequisites.
  3. Any 'focus on' sub-goal appears after the sub-goal that finds or obtains the object.
  4. If it is the 4th (last) sub-goal, completing it achieves the final goal stated in the Task.
  5. Sub-goals use generic descriptions for variable elements (e.g., 'the room where X can be found').
Rate how well the Partial plan satisfies the criteria above on a continuous scale from 0.0 to 1.0 (1.0 = fully correct and well-ordered, 0.0 = clearly wrong). Use the full range: a plan that satisfies most but not all criteria should receive an intermediate value.
Answer with a single decimal number only (e.g., 0.82). No explanation.
\end{lstlisting}
\end{promptbox}

\begin{promptbox}{Beam Search Planner in-context example}
\small
\begin{lstlisting}
USER: Task: Your task is to use chemistry to create green paint. When you are done, focus on the green paint.
Current plan so far: []
Generate the next sub-goal for THIS task only.

ASSISTANT: Teleport to the room where the needed materials are found

USER: Task: Your task is to use chemistry to create green paint. When you are done, focus on the green paint.
Current plan so far: ['Teleport to the room where the needed materials are found']
Generate the next sub-goal for THIS task only.

ASSISTANT: Gather the materials needed for green paint

USER: Task: Your task is to use chemistry to create green paint. When you are done, focus on the green paint.
Current plan so far: ['Teleport to the room where the needed materials are found', 'Gather the materials needed for green paint']
Generate the next sub-goal for THIS task only.

ASSISTANT: Mix the materials to create green paint

USER: Task: Your task is to use chemistry to create green paint. When you are done, focus on the green paint.
Current plan so far: ['Teleport to the room where the needed materials are found', 'Gather the materials needed for green paint', 'Mix the materials to create green paint']
Generate the next sub-goal for THIS task only.

ASSISTANT: Focus on the resulting green paint
\end{lstlisting}
\end{promptbox}

\paragraph{State-change planner prompts (+proc).}\mbox{}

The \emph{+proc} condition of Table~\ref{tab:procedure} inserts the rule below
into both planner prompts above as rule (3), moving each prompt's
generic-description rule to (4). The rule is identical in both.

\begin{promptbox}{Added rule for the \emph{+proc} condition}
\small
\begin{lstlisting}
(3) ONLY for tasks explicitly involving boiling, melting, freezing, or changing the state of matter of a substance -- NOT for chemistry, mixing, or paint creation tasks -- NEVER use 'take actions to change state of matter' as a sub-goal -- it is too vague. The correct sub-goal order is: (a) find and obtain the substance, (b) focus on the substance, (c) reach or obtain the heat or cold source, (d) apply heat or cold to cause the state change. The 'focus on the substance' sub-goal MUST come at step (b) -- never as the last sub-goal.
\end{lstlisting}
\end{promptbox}

For the Beam Search Planner that is the only change. The single-path variant
additionally takes rules (1) and (2) from the Beam Search Planner prompt,
dropping the two phrases that apply only during incremental expansion:
\texttt{the next} in rule (1), and \texttt{when the current plan is empty} in
rule (2).

\section{Jericho Planner Prompts and In-Context Example}
\label{app:jericho_planner}

\begin{promptbox}{One-shot planner prompt}
\small
\begin{lstlisting}
Generate exactly 4 high-level sub-goals for a task in a text-based interactive fiction adventure game (Jericho). Each sub-goal is an outcome or milestone -- WHAT to achieve, not HOW -- and may require several actions including exploration and navigation. Rules: (1) Order sub-goals by dependency: obtain an item before using or giving it; if the goal mentions darkness, obtain a light source before entering dark areas. (2) Use the names and hints stated in the goal text (a specific room, where an item is hidden, a required tool or method); for anything not stated, use a generic description and never invent specific items, locations, or methods. (3) The final sub-goal must directly achieve the objective (obtain X, give X to Y, or reach the exit); add no steps after it. Format: 'Sub goals:\n1. ...\n2. ...\n...'
\end{lstlisting}
\end{promptbox}

\begin{promptbox}{One-shot planner in-context example}
\small
\begin{lstlisting}
USER: Goal: You are the warrior Link that needs to save the princess from the castle.

Starting room:
You are at the path leading to the castle. The castle is to your north. There is a barrel in front of you.

ASSISTANT: Sub goals:
1. Find and obtain any useful item available in the starting area
2. Navigate into the castle
3. Reach the princess's location within the castle
4. Wake the princess to complete the rescue
\end{lstlisting}
\end{promptbox}

\begin{promptbox}{Beam Search Planner prompt (per-step candidate generation)}
\small
\begin{lstlisting}
Generate the next high-level sub-goal to extend a partial plan for a task in a text-based interactive fiction adventure game (Jericho). Each sub-goal is an outcome or milestone -- WHAT to achieve, not HOW -- and may require several actions including exploration and navigation. Rules: (1) Derive the sub-goal from the task goal and starting observation, not from the example. (2) Order by dependency: obtain an item or light source before using it or entering dark areas; do not depend on something a later sub-goal will provide. (3) Use the names and hints stated in the goal text; for anything not stated, use a generic description and never invent specific items, locations, or methods. (4) If this is the 4th (last) sub-goal, it must directly achieve the objective (obtain X, give X to Y, or reach the exit). The total plan must consist of exactly 4 sub-goals. Output ONLY the sub-goal description on a single line. No explanation, no prefix.
\end{lstlisting}
\end{promptbox}

Here \texttt{\{expand\_prompt\}} is the Beam Search Planner prompt above.

\begin{promptbox}{Self-evaluation prompt}
\small
\begin{lstlisting}
Evaluate whether a partial plan is appropriate given the following planning constraints:

{expand_prompt}

Given the Task (goal + starting room) and its Partial plan, assess how well the plan satisfies ALL of the following:
  1. Each sub-goal describes WHAT to achieve (an outcome or milestone), not HOW.
  2. No sub-goal contains a bare navigation direction (north, south, east, west, up, down) as its main action -- destinations must be described by what they contain or their purpose instead.
  3. Each sub-goal logically follows from the previous ones without skipping prerequisites.
  4. An item must be found or obtained before any sub-goal that uses or gives it.
  5. If the goal mentions darkness or a required light source, the sub-goal to obtain the light source appears before any sub-goal that enters a dark area.
  6. Sub-goals use generic descriptions for items or locations not explicitly stated in the goal or starting observation (e.g., 'the required tool', 'the target location') rather than hallucinating specific names.
  7. If it is the 4th (last) sub-goal, completing it directly achieves the final objective stated in the Task (e.g., 'deliver X to Y', 'open the gate'), not merely gather information.
Rate how well the Partial plan satisfies the criteria above on a continuous scale from 0.0 to 1.0 (1.0 = fully correct and well-ordered, 0.0 = clearly wrong). Use the full range: a plan that satisfies most but not all criteria should receive an intermediate value.
Answer with a single decimal number only (e.g., 0.82). No explanation.
\end{lstlisting}
\end{promptbox}

\begin{promptbox}{Beam Search Planner in-context example}
\small
\begin{lstlisting}
USER: Task: Goal: You are the warrior Link that needs to save the princess from the castle.

Starting room:
You are at the path leading to the castle. The castle is to your north. There is a barrel in front of you.
Current plan so far: []
Generate the next sub-goal (step 1 of 4):

ASSISTANT: Find and obtain any useful item available in the starting area

USER: Task: Goal: You are the warrior Link that needs to save the princess from the castle.

Starting room:
You are at the path leading to the castle. The castle is to your north. There is a barrel in front of you.
Current plan so far: ['Find and obtain any useful item available in the starting area']
Generate the next sub-goal (step 2 of 4):

ASSISTANT: Navigate into the castle

USER: Task: Goal: You are the warrior Link that needs to save the princess from the castle.

Starting room:
You are at the path leading to the castle. The castle is to your north. There is a barrel in front of you.
Current plan so far: ['Find and obtain any useful item available in the starting area', 'Navigate into the castle']
Generate the next sub-goal (step 3 of 4):

ASSISTANT: Reach the princess's location within the castle

USER: Task: Goal: You are the warrior Link that needs to save the princess from the castle.

Starting room:
You are at the path leading to the castle. The castle is to your north. There is a barrel in front of you.
Current plan so far: ['Find and obtain any useful item available in the starting area', 'Navigate into the castle', "Reach the princess's location within the castle"]
Generate the next sub-goal (step 4 of 4):

ASSISTANT: Wake the princess to complete the rescue
\end{lstlisting}
\end{promptbox}

\clearpage
\section{Ablation, Grounding, and Procedural-Prompting Figures/Tables}
\label{app:moved_tables}

\begin{figure}[h]
\centering
\begin{tikzpicture}
\begin{axis}[
  width=\textwidth, height=4.1cm,
  ybar, bar width=4.6pt, ybar legend,
  ymin=0, ymax=82, ytick={0,20,40,60,80},
  ylabel={Success rate (\%)}, ylabel style={font=\small, yshift=-4pt},
  xmin=0.45, xmax=11.55,
  xtick={1,2,3,4,5,6,7,8,9,10,11},
  xticklabels={%
    {Boil~{\scriptsize(n\,{=}\,9)}},{Freeze~{\scriptsize(n\,{=}\,9)}},
    {Change-State~{\scriptsize(n\,{=}\,9)}},{Melt~{\scriptsize(n\,{=}\,9)}},
    {Paint\,/\,Mix~{\scriptsize(n\,{=}\,18)}},{Find~{\scriptsize(n\,{=}\,40)}},
    {Lifespan~{\scriptsize(n\,{=}\,30)}},{Measure~{\scriptsize(n\,{=}\,20)}},
    {Grow~{\scriptsize(n\,{=}\,20)}},{Other~{\scriptsize(n\,{=}\,47)}},
    {\textbf{Total}~{\scriptsize(n\,{=}\,211)}}},
  xticklabel style={rotate=60, anchor=east, align=right, font=\small,
                    yshift=1pt, xshift=2pt},
  xtick style={draw=none},
  ymajorgrids=true, grid style={draw=black!12, line width=0.3pt},
  axis x line*=bottom, axis y line*=left,
  tick label style={font=\small},
  legend style={at={(0.5,-0.54)}, anchor=north, legend columns=4,
                draw=black!25, font=\small, column sep=6pt},
  clip=false,
]
\fill[black!5] (axis cs:0.5,0) rectangle (axis cs:4.5,82);
\draw[black!12, line width=0.3pt] (axis cs:0.5,20) -- (axis cs:4.5,20);
\draw[black!12, line width=0.3pt] (axis cs:0.5,40) -- (axis cs:4.5,40);
\draw[black!12, line width=0.3pt] (axis cs:0.5,60) -- (axis cs:4.5,60);
\draw[black!12, line width=0.3pt] (axis cs:0.5,80) -- (axis cs:4.5,80);
\node[anchor=south, font=\small, text=black!65]
  at (axis cs:2.5,74) {State-change tasks};
\draw[black!25, line width=0.3pt] (axis cs:1.5,0) -- (axis cs:1.5,82);
\draw[black!25, line width=0.3pt] (axis cs:2.5,0) -- (axis cs:2.5,82);
\draw[black!25, line width=0.3pt] (axis cs:3.5,0) -- (axis cs:3.5,82);
\draw[black!25, line width=0.3pt] (axis cs:4.5,0) -- (axis cs:4.5,82);
\draw[black!25, line width=0.3pt] (axis cs:5.5,0) -- (axis cs:5.5,82);
\draw[black!25, line width=0.3pt] (axis cs:6.5,0) -- (axis cs:6.5,82);
\draw[black!25, line width=0.3pt] (axis cs:7.5,0) -- (axis cs:7.5,82);
\draw[black!25, line width=0.3pt] (axis cs:8.5,0) -- (axis cs:8.5,82);
\draw[black!25, line width=0.3pt] (axis cs:9.5,0) -- (axis cs:9.5,82);
\draw[dashed, black!40] (axis cs:10.5,0) -- (axis cs:10.5,82);

\addplot[draw=black!55, fill=cReAct] coordinates
  {(1,0.0) (2,0.0) (3,11.1) (4,0.0) (5,22.2) (6,45.0) (7,40.0) (8,0.0) (9,5.0) (10,6.4) (11,18.5)};
\addplot[draw=black!65, fill=cReflAct] coordinates
  {(1,0.0) (2,0.0) (3,11.1) (4,33.3) (5,5.6) (6,70.0) (7,66.7) (8,0.0) (9,5.0) (10,0.0) (11,25.6)};
\addplot[draw=cTeal, pattern=north east lines, pattern color=cTeal] coordinates
  {(1,22.2) (2,0.0) (3,33.3) (4,11.1) (5,0.0) (6,70.0) (7,53.3) (8,10.0) (9,0.0) (10,8.5) (11,26.5)};
\addplot[draw=cTeal!80!black, fill=cTeal] coordinates
  {(1,55.6) (2,0.0) (3,33.3) (4,0.0) (5,61.1) (6,77.5) (7,53.3) (8,20.0) (9,5.0) (10,0.0) (11,33.6)};

\legend{\react{}, \reflact{}, \beammethod{}, \method{} (ours)}
\end{axis}
\end{tikzpicture}
\caption{
  ScienceWorld success rates by task bucket across four reasoning backbones
  (Llama-3.1-8B-Instruct). The ten buckets regroup ScienceWorld's nine
  categories for readability: the state-change category is shown as its four
  task types (shaded), the chemistry category as its paint-mixing episodes,
  and the rest are pooled into ``Other'' (electrical/power, conductivity
  testing, non-paint chemistry mixing, and life-stage identification). $n$ is
  the number of episodes per bucket, and ``Total'' is the rate over 211
  episodes, matching Table~\ref{tab:main}.
  Table~\ref{tab:tasktype11} gives the per-row values behind these bars, and
  Table~\ref{tab:scicat} the same episodes by the nine categories of the
  ScienceWorld split.
}
\label{fig:tasktype}
\end{figure}
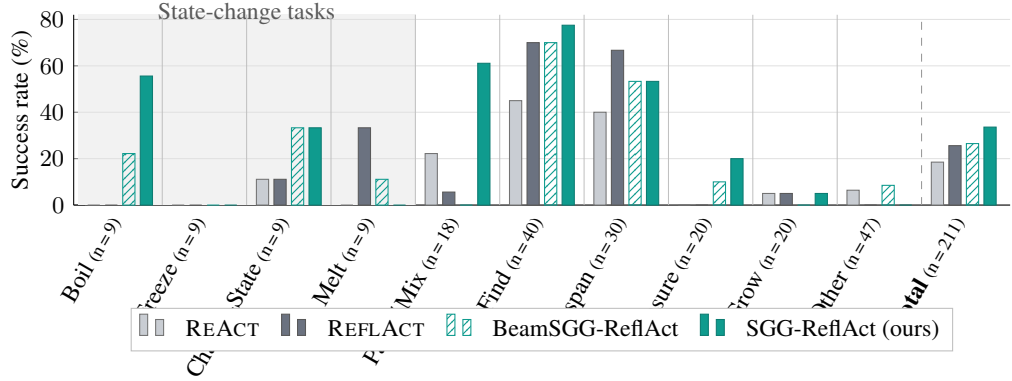

Table~\ref{tab:ablation} accompanies the reflection-anchoring ablation in
Section~\ref{sec:task_analysis}; Figure~\ref{fig:compounding} and
Figure~\ref{fig:tasktype} accompany the grounding analysis in the same
section; and Table~\ref{tab:procedure} accompanies the
procedural-prompting analysis in Section~\ref{sec:plan_quality}.

\begin{figure}[h]
\centering
\begin{tikzpicture}
\begin{axis}[
  width=\linewidth, height=4.4cm,
  xlabel={Episode position}, ylabel={Invalid-action rate (\%)},
  xlabel style={font=\small}, ylabel style={font=\small, yshift=-3pt},
  xtick={1,2,3}, xticklabels={early, mid, late},
  ymin=0, ymax=80, ytick={0,20,40,60,80},
  xmin=0.88, xmax=3.12,
  ymajorgrids=true, grid style={draw=black!12, line width=0.3pt},
  axis x line*=bottom, axis y line*=left,
  tick label style={font=\small},
  legend style={at={(0.5,-0.32)}, anchor=north, legend columns=2,
                draw=black!25, font=\scriptsize, column sep=4pt,
                /tikz/every even column/.append style={column sep=4pt}},
  legend cell align=left,
  every mark/.append style={scale=0.75},
]
\addplot[color=cNoTh, dotted, line width=0.9pt, mark=diamond*,
         mark options={solid, fill=cNoTh, draw=black!45}]
  coordinates {(1,20.4) (2,43.8) (3,72.2)}; \addlegendentry{NoThinking}
\addplot[color=cPlanAct, dashdotted, line width=0.9pt, mark=pentagon*,
         mark options={solid, fill=cPlanAct, draw=black!45}]
  coordinates {(1,17.2) (2,39.0) (3,62.4)}; \addlegendentry{Plan-and-Act}
\addplot[color=cReActC, densely dashed, line width=0.9pt, mark=o,
         mark options={solid, draw=cReActC}]
  coordinates {(1,18.6) (2,47.4) (3,68.7)}; \addlegendentry{\react{}}
\addplot[color=cReflAct, line width=0.9pt, mark=*,
         mark options={solid, fill=cReflAct, draw=cReflAct}]
  coordinates {(1,14.5) (2,44.1) (3,71.4)}; \addlegendentry{\reflact{}}
\addplot[color=cTeal, densely dashed, line width=1.0pt, mark=triangle*,
         mark options={solid, fill=white, draw=cTeal}]
  coordinates {(1,9.3) (2,36.6) (3,56.9)}; \addlegendentry{\beammethod{}}
\addplot[color=cTeal, line width=1.1pt, mark=square*,
         mark options={solid, fill=cTeal, draw=cTeal}]
  coordinates {(1,8.7) (2,31.2) (3,49.7)}; \addlegendentry{\method{}}
\end{axis}
\end{tikzpicture}
\caption{
  Invalid-action rate by episode third on ScienceWorld
  (Llama-3.1-8B-Instruct) for all six agents; see
  Appendix~\ref{app:grounding_table} for GPT-4o and GPT-4o-mini results.
}
\label{fig:compounding}
\end{figure}
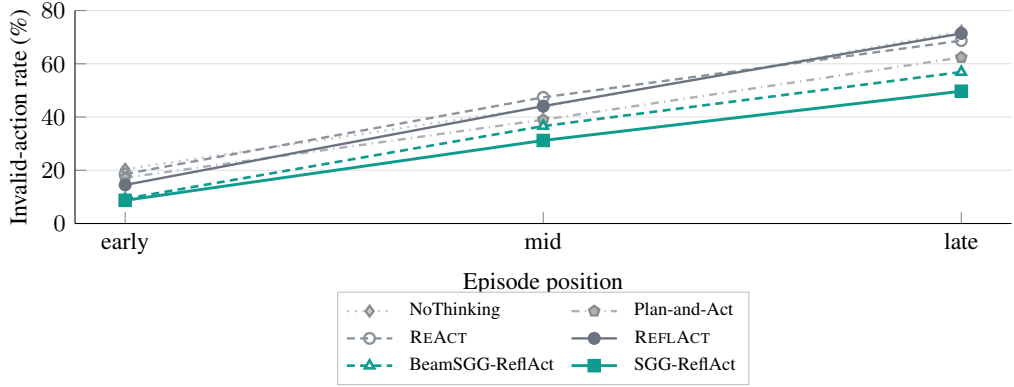

\begin{table}[h]
\centering
\caption{
  Ablation isolating the reflection-anchoring instruction on ScienceWorld
  (Llama-3.1-8B-Instruct). \emph{Plan} indicates whether a sub-goal plan is
  present in the context. Values are percentages.
}
\label{tab:ablation}
\footnotesize
\setlength{\tabcolsep}{6pt}
\begin{tabular}{lccc}
\toprule
Agent & Plan & Reflects on & SR / AR \\
\midrule
\reflact{}        & \texttimes & end-goal & 25.6 / 41.5 \\
\reflact{}~+~Plan & \checkmark & end-goal & 27.5 / 44.8 \\
\method{}         & \checkmark & active sub-goal & \textbf{33.6} / \textbf{48.7} \\
\bottomrule
\end{tabular}
\end{table}

\begin{table}[h]
\centering
\caption{
  ScienceWorld success rates (\%) for both planners, comparing the base
  ScienceWorld planner prompt with its state-change variant
  (\emph{+proc}; Appendix~\ref{app:sci_planner}). \emph{Overall} includes 211
  episodes; \emph{State-change} includes the 36 targeted episodes
  (Boil/Freeze/Melt/Change-State). SGG/Beam denote \method{}/\beammethod{}.
  \emph{base} matches Table~\ref{tab:main}. Bold indicates the best
  performance in each pair.
}
\label{tab:procedure}
\small
\setlength{\tabcolsep}{5pt}
\begin{tabular}{llcccc}
\toprule
 & & \multicolumn{2}{c}{Overall SR} & \multicolumn{2}{c}{State-change SR} \\
\cline{3-4}\cline{5-6}
Model & Planner & base & +proc & base & +proc \\
\midrule
\multirow{2}{*}{GPT-4o} & SGG & 64.0 & \textbf{66.4} & 55.6 & \textbf{72.2} \\
 & Beam & 63.5 & \textbf{65.4} & 66.7 & \textbf{75.0} \\
\midrule
\multirow{2}{*}{GPT-4o-mini} & SGG & 45.0 & \textbf{49.8} & 52.8 & \textbf{55.6} \\
 & Beam & 43.6 & \textbf{47.4} & 25.0 & \textbf{55.6} \\
\midrule
\multirow{2}{*}{\shortstack[l]{Llama-3.1\\-8B}} & SGG & \textbf{33.6} & 32.2 & \textbf{22.2} & 16.7 \\
 & Beam & 26.5 & \textbf{31.3} & 16.7 & \textbf{27.8} \\
\bottomrule
\end{tabular}
\end{table}

\end{document}